\documentclass[11pt]{article}

\usepackage[T1]{fontenc}

\usepackage[margin=1in]{geometry}

\usepackage[american]{babel}

\usepackage{natbib} % has a nice set of citation styles and commands
\usepackage{xcolor}
\usepackage{booktabs}
\usepackage{amsmath}
\usepackage{amssymb}
\usepackage{mathtools}
\usepackage{siunitx}
\usepackage{tikz} % nice language for creating drawings and diagrams
\usepackage{subcaption}
\usetikzlibrary{arrows.meta, positioning, fit, backgrounds}
\usetikzlibrary{bayesnet}
\tikzset{plate caption/.append style={font=\small, below left=0pt and 0pt of #1.south east}}
\usepackage{lipsum} % For filler text
\tikzset{
    bayesnode/.style={draw, circle, text width=0.4cm, align=center, minimum size=0.55cm, font=\small},
    latent/.style={bayesnode, fill=white},
    observed/.style={bayesnode, fill=gray!30, text width = 0.5cm},
    param/.style={draw, rectangle, text width=0.55cm, fill=gray!20, minimum size=0.75cm, aspect=1, align=center, font=\small},
    edge/.style={-Stealth},
    dashededge/.style={-Stealth,dashed}
}
\usepackage{tabularx}    % for automatic column width adjustment
\usepackage{multirow}    % for row grouping if needed
\usepackage{makecell}
\usepackage{adjustbox}
\usepackage{array}

\renewcommand\appendix{\par 
    \setcounter{section}{0}%   
    \setcounter{subsection}{0}%  
    \setcounter{subsubsection}{0}%  
    \setcounter{table}{0} 
    \setcounter{figure}{0}
    \gdef\thefigure{\Alph{section}.\arabic{figure}}%
    \gdef\thetable{\Alph{section}.\arabic{table}}%
    \gdef\thesection{\Alph{section}}%
    \gdef\thesubsection{\Alph{section}.\arabic{subsection}}% 
    \gdef\theequation{\Alph{section}.\arabic{equation}}%
}
\usepackage{etoolbox}
\newtoggle{inappendix}
\togglefalse{inappendix}
\apptocmd{\appendix}{\toggletrue{inappendix}}{}{\errmessage{failed to patch}}

\newcommand{\E}{\mathbb{E}} % expectation
\newcommand{\g}{\,|\,} % given
\newcommand{\p}{\textrm{Pr}} % probability
\newcommand{\reals}{\mathbb{R}} % real line
\newcommand\inv[1]{#1\raisebox{1.15ex}{$\scriptscriptstyle-\!1$}}

\newcommand{\distas}{\quad\sim\quad}

\newcommand{\indep}{\perp \!\!\! \perp}

\newcommand{\GP}{\ensuremath{\textrm{GP}}}
\newcommand{\obs}{{\textrm{obs}}}

\newcommand{\bfzero}{\mathbf{0}}

\newcommand{\bfbeta}{{\boldsymbol{\beta}}}
\newcommand{\bfgamma}{{\boldsymbol{\gamma}}}

\newcommand{\bfzeta}{{\boldsymbol{\zeta}}}

\newcommand{\bftheta}{{\boldsymbol{\theta}}}

\newcommand{\bflambda}{{\boldsymbol{\lambda}}}
\newcommand{\bfmu}{{\boldsymbol{\mu}}}

\newcommand{\bfpi}{{\boldsymbol{\pi}}}
\newcommand{\bfrho}{{\boldsymbol{\rho}}}

\newcommand{\bftau}{{\boldsymbol{\tau}}}

\newcommand{\bfphi}{{\boldsymbol{\phi}}}

\newcommand{\bfTheta}{\boldsymbol{\Theta}}

\newcommand{\bfm}{\mathbf{m}}

\newcommand{\bfx}{\mathbf{x}}

\newcommand{\bfK}{\mathbf{K}}

\newcommand{\bfX}{\mathbf{X}}

\newcommand{\calD}{\mathcal{D}}

\newcommand{\bernoulli}{\textrm{Bern}}
\newcommand{\betarand}{\mathrm{Beta}}

\newcommand{\categorical}{\mathrm{Cat}}

\newcommand{\normal}{\mathcal{N}}

\newcommand{\logit}{\mathrm{logit}}

\newcommand{\ticks}{$\checkmark$}
\newcommand{\cross}{$\times$}
\newcommand{\basiccs}{\textsc{bscc}}
\newcommand{\simHTE}{\textsc{simulation hte}}
\newcommand{\simFS}{\textsc{simulation fs}}
\newcommand{\simNull}{\textsc{simulation null}}

\usepackage{hyperref}
\hypersetup{colorlinks=true, linkcolor=blue, citecolor=blue, urlcolor=blue}

\makeatletter
\patchcmd{\hyper@makecurrent}{%
    \ifx\Hy@param\Hy@chapterstring
        \let\Hy@param\Hy@chapapp
    \fi
}{%
    \iftoggle{inappendix}{%true-branch
        \@checkappendixparam{chapter}%
        \@checkappendixparam{section}%
        \@checkappendixparam{subsection}%
        \@checkappendixparam{subsubsection}%
        \@checkappendixparam{paragraph}%
        \@checkappendixparam{subparagraph}%
    }{}%
}{}{\errmessage{failed to patch}}
\newcommand*{\@checkappendixparam}[1]{%
    \def\@checkappendixparamtmp{#1}%
    \ifx\Hy@param\@checkappendixparamtmp
        \let\Hy@param\Hy@appendixstring
    \fi
}
\makeatother
\title{\textbf{Bayesian Supervised Causal Clustering}}

\author{
    Luwei~Wang\textsuperscript{1} \and
    Nazir~Lone\textsuperscript{2} \and
    Sohan~Seth\textsuperscript{1} \\[6pt]
    \textsuperscript{1}School of Informatics, University of Edinburgh, Edinburgh, UK \\
    \textsuperscript{2}Usher Institute, University of Edinburgh, Edinburgh, UK
}
\date{}

\begin{document}
\maketitle

\begin{abstract}
Finding patient subgroups with similar characteristics is crucial for personalized decision-making in various disciplines such as healthcare and policy evaluation. While most existing approaches rely on unsupervised clustering methods, there is a growing trend toward using supervised clustering methods that identify operationalizable subgroups in the context of a specific outcome of interest. 
We propose Bayesian Supervised Causal Clustering ({\basiccs}), with \emph{treatment effect as outcome} to guide the clustering process. % under the assumptions of randomization.
{\basiccs} identifies homogenous subgroups of individuals who are similar in their covariate profiles as well as their treatment effects. 
We evaluate {\basiccs} on simulated datasets %to demonstrate its ability in handling multiple mixed covariates and nonlinear relationships between covariates and potential outcomes.
% We also apply {\basiccs} to a 
as well as real-world dataset from the third International Stroke Trial to assess the practical usefulness of the framework. %The code is available at [github.com].
\end{abstract}

\section{Introduction}\label{sec:intro}

Patient stratification, i.e., the identification of subgroups sharing clinical, molecular, or genetic characteristics, is a cornerstone of precision medicine and data-driven decision-making \citep{Abdelnour2022PS}. By uncovering structured variation across high-dimensional population data, stratification enables clinicians to tailor treatments to specific groups, maximizing efficacy while minimizing adverse effects \citep{Dubois2015PS}. Crucially, this approach addresses the challenge of Heterogeneity of Treatment Effects (HTE), where identical interventions yield divergent responses across subpopulations due to complex interactions between patient characteristics and treatment mechanisms \citep{Thelen2025}. While Randomized Controlled Trials (RCTs) generally report Average Treatment Effects (ATE), such aggregate summaries often mask substantial variability in treatment response \citep{Shehabi2021}. Ignoring this heterogeneity can lead to ineffective or even harmful clinical decisions: a modest overall benefit might obscure a large effect in a small, identifiable subgroup, or conversely, significant harm in another \citep{Kravitz2004HTE, Kent2018HTE-example}. Consequently, identifying \emph{clinically operationalizable} subgroups that are distinct not only in their observable features but also in their response to treatment is essential for guiding personalized therapeutic strategies.

\newcommand{\gmm}{\textsc{gmm}}
\newcommand{\lca}{\textsc{lca}}

\begin{figure*}[!htb]
    \centering
    \includegraphics[width=\linewidth]{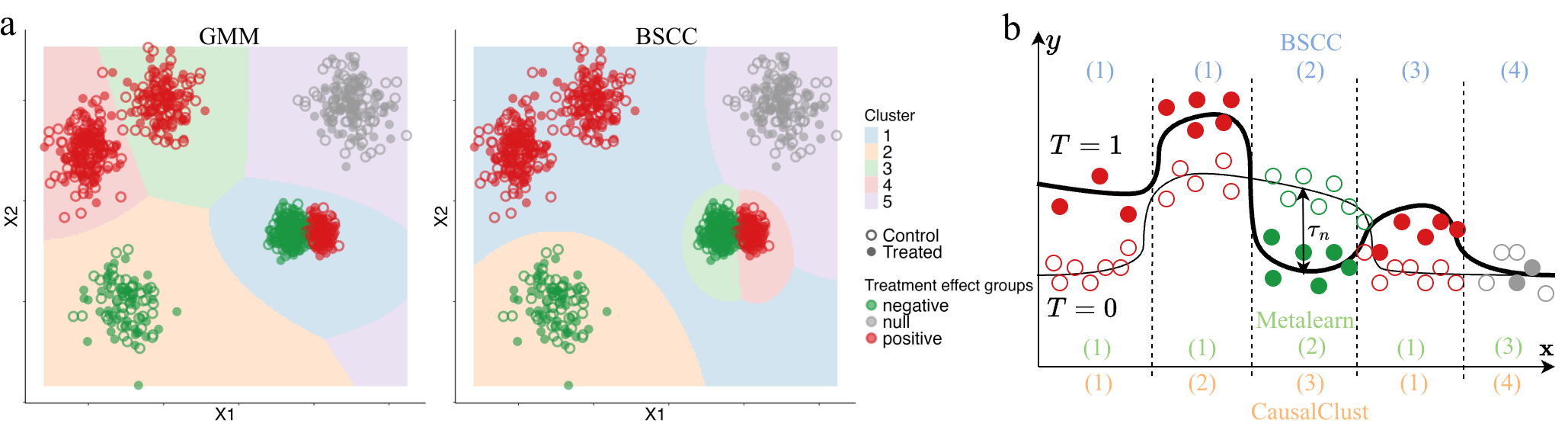}
    \caption{\textbf{(a)} Illustrating the conceptual difference between \textbf{(left)} unsupervised clustering with Gaussian Mixture Model ({\gmm}) \textbf{(right)} supervised clustering with {\basiccs}. {\gmm} fails to recover the subgroups relevant to treatment while {\basiccs} can. \textbf{(b)} Difference between {\basiccs} clustering samples based on both covariates and treatment effect in a supervised clustering set-up, as opposed to causal clustering or supervised learning based approaches clustering based on potential outcomes or treatment effect only. }
    \label{fig:demo-simulated_data2D_byGMM&HTEclustX}
\end{figure*}

% Clustering
Traditionally, unsupervised clustering methods such as Gaussian Mixture Models (\textbf{\gmm}) \citep{ChrisGMM2002} and Latent Class Analysis (\textbf{\lca}) \citep{CollinsLCA2010} have been the primary tools for data-driven patient stratification. These methods excel at uncovering hidden population structures by minimizing within-group variance in the covariate space \citep{MurphyML2012}. Approaches like $k$-means and {\lca} have been successfully applied to characterize disease phenotypes in conditions such as sepsis \citep{Seymour2019KmeansExample}, acute respiratory distress syndrome (ARDS) \citep{Sinha2018LDAexample}, and hypertension \citep{Guo2017HypertensionExm}. However, a key limitation of unsupervised clustering is that the resulting subgroups are defined solely by covariate similarity, ignoring the outcome or treatment response. As a consequence, the identified clusters may be phenotypically distinct but heterogeneous with respect to treatment effects, limiting their utility for prescriptive decision-making (Figure~\ref{fig:demo-simulated_data2D_byGMM&HTEclustX}\textbf{(a, left)}).

To address this limitation, recent methodological research has shifted toward supervised clustering, which incorporates outcome information into the clustering process to ensure that the discovered groups are predictive of the target variable \citep{Rouanet2023BayesianProfile}. However, most existing supervised clustering methods focus on predicting an outcome rather than on predicting the difference between two potential outcomes. In the context of precision medicine, it might also be beneficial to identify groups with different responses to an intervention (i.e., predictive enrichment) alongside those with different outcomes (i.e., prognostic enrichment) \citep{Temple2010ProgPredEnrich}. This motivates the development of \emph{supervised causal clustering}, where subgroup discovery is guided jointly by covariate similarity and treatment-effect variation. Such a framework offers a promising path toward discovering subgroups that are both interpretable in terms of patient characteristics and actionable in terms of distinct treatment responses.

\paragraph{Related Work}
We summarize existing approaches for subgroup discovery in Table~\ref{tab:related_work}. %, organized by clustering target and supervision type.
\emph{Unsupervised clustering} (\textsc{UC}) methods cluster patients based on covariate similarity alone, ignoring treatment response \citep{ChrisGMM2002,CollinsLCA2010}. While useful for exploratory analysis, they do not guarantee that the resulting clusters exhibit homogeneous treatment effects.

\newcommand{\sgmm}{\textsc{sgmm}}

\emph{Supervised clustering with outcome} (\textsc{SC}) as side information, including Bayesian profile regression \citep{Molitor2010BayesProfile} and supervised GMM (\textbf{\sgmm}) \citep{Shou2019supervisedMixture,lyu2024semisupervisedGMM}, jointly model covariates for clustering while predicting outcomes via fixed effects. Extensions accommodate various outcome types, e.g., categorical and count data \citep{Liverani2015FeatureSelection} and longitudinal outcomes \citep{Rouanet2023BayesianProfile}. However, treatment effect has not been the primary focus of these studies, and they typically model the conditional expectation of the outcome rather than the contrast between potential outcomes.

\newcommand{\mob}{\textsc{mob}}
\newcommand{\itt}{\textsc{it}}

\emph{Tree-based subgroup analysis} (\textsc{SA}) methods, including interaction trees (\textbf{\itt}) \citep{Su2008InteractionTree} and model-based recursive partitioning (\textbf{\mob}) \citep{Zeileis2008MOB,Seibold2016MoB}, partition the covariate space recursively based on treatment effect heterogeneity. While directly supervised by treatment effect, these greedy and deterministic approaches may miss subgroups defined by complex variable interactions or soft boundaries, and can be unstable \citep{Thelen2025}.

\newcommand{\cf}{\textsc{cf}}
\newcommand{\cc}{\textsc{cc}}
\newcommand{\bart}{\textsc{bart}}
\newcommand{\findit}{\textsc{FindIt}}
\newcommand{\owe}{\textsc{owe}}

\emph{Effect modeling} (\textsc{EM}) methods estimate individual treatment effects (ITE) using meta-learners \citep{Kennedy2023}: doubly robust learner (\textbf{\textsc{DR-learner}}) \citep{Kennedy2020DRlearner} and residual-learner (\textbf{\textsc{R-learner}})) \citep{Nie2020Rlearner}; causal forests (\textbf{\cf}) \citep{Athey2019causalForest}, Bayesian additive regression trees (\textbf{\bart}) \citep{Sparapani2021BART}, or penalized regressions like Finding Heterogeneous Treatment Effects (\textbf{\findit}) \citep{Imai2013FindIt} and Outcome Weighted Estimation (\textbf{\owe}) \citep{Shuai2017OWE}. While these methods provide flexible effect estimates at an individual level, they lack interpretable subgroup structures. Recovering subgroups from ITE estimates typically requires a post-hoc clustering step, %which may not align with the original data structure 
\citep{bonvini2023}. 
\emph{Causal clustering} (\textbf{\cc}) \citep{Kim2024HierCausalClust} clusters individuals based on their potential outcomes estimated via counterfactual regression. This approach disentangles the potential outcome structure but ignores the covariate structure, potentially conflating structurally distinct subpopulations that happen to share similar potential outcomes (Figure~\ref{fig:demo-simulated_data2D_byGMM&HTEclustX}\textbf{(b)}).

\emph{Bayesian methods} (\textsc{BM}) have been used to capture HTE without focusing on finding subgroups in the covariate space.
\citet{Xu2016BNP} use mixture modelling to group longitudinal treatment effects, and \citet{Shahn2017LatentClass} captures treatment heterogeneity as a mixture model. %propose latent class mixture models that partition individuals into binary classes with distinct treatment effects, augmented by regression interaction terms for continuous heterogeneity. 
%While both approaches provide principled Bayesian uncertainty quantification, neither jointly models covariate similarity and treatment effect for subgroup discovery: the former focuses on individual response curve estimation without explicit subgroup structure, and the latter defines latent classes primarily through treatment response patterns rather than covariate-driven clusters.

\begin{table}[!t]
\centering
\caption{Summary of approaches. \textbf{TE}: treatment-effect-based.}
\label{tab:related_work}
\footnotesize
\setlength{\tabcolsep}{4pt}
\begin{tabular}{@{}lccccl@{}}
\toprule
\textbf{Category} & \textbf{Supervised} & \textbf{Probabilistic} & \textbf{Covariate-based} & \textbf{TE} & \textbf{Example} \\
\midrule
\textsc{UC} & \cross & \ticks & \ticks & \cross & {\gmm} \\\hline
\addlinespace
\textsc{SC} & \ticks & \ticks & \ticks & \cross & {\sgmm} \\\hline
\addlinespace
\multirow{2}{*}{\textsc{SA}} & \ticks & \cross & \ticks & \ticks & {\itt}, {\mob} \\
 & \ticks & \ticks & \ticks & \ticks & \textbf{{\basiccs} (ours)} \\\hline
\addlinespace
\textsc{EM} & \ticks & \cross & \cross & \ticks & {\cf}, {\bart}, \textsc{DR-l}, \textsc{R-l}, {\findit}, {\owe}, {\cc} \\\hline
\addlinespace
\textsc{BM} & \ticks & \ticks & \cross & \ticks & \citet{Xu2016BNP}, \citet{Shahn2017LatentClass} \\
\bottomrule
\end{tabular}
\end{table}
%In contrast, our proposed method clusters based on \emph{both} covariate similarity and treatment effect, combining the ease of interpretation of covariate-based clustering with the relevance of causal effect supervision.

\paragraph{Contributions}
% advantages: 1. causal effect: treatment effect. 2. clustering
We propose Bayesian Supervised Causal Clustering, or \textbf{\basiccs} (``basic''), as a probabilistic framework for operationalizable patient stratification. {\basiccs} jointly clusters individuals based on both their covariate structure and their response to treatment. Unlike traditional unsupervised clustering methods that only consider covariate similarity, or traditional effect modelling methods that focus on ITE estimation without finding subgroups, or causal clustering methods that do not consider covariate structure, {\basiccs} explicitly integrates causal effect as a supervisory signal into the clustering process. Similar to tree-based subgroup analysis, this allows for the identification of subgroups that are homogeneous in covariates and treatment effects, but with more flexible subgroups through Bayesian mixture modelling. We explore the effectiveness of {\basiccs} in both simulated and real datasets extensively, demonstrating that it can identify meaningful and consistent subgroups of individuals.

\section{Methods}\label{sec:method}

Standard mixture model with $K$ components is described as 
%{\small
$$ 
p(\bfx_n)=\sum_{k=1}^K p(z_n=k)p(\bfx_n\g z_n=k)=\sum_{k=1}^K \pi_k p(\bfx_n \g \bftheta_k)
$$ %}
where $\bfx_n\in\reals^D$ is the covariate vector for individual $n$, $\pi_k$ is the probability of the $k$-th mixture component and $\bftheta_k$ is the parameter of the respective component. $z_n$ is the latent cluster assignment for the individual. When incorporating the information from outcome variable $y$, the supervised mixture model \citep{Shou2019supervisedMixture} is described as 
%{\small
\begin{align*}
p(y_n,\bfx_n) %&= \sum_{k=1}^K p(y_n\g\bfx_n, z_n=k)p(\bfx_n\g z_n=k)p(z_n=k)\\&
=\sum_{k=1}^K \pi_kp(\bfx_n \g \bftheta_k) p(y_n\g \bfx_n, \bfbeta_k)
\end{align*} %}
where $\bfbeta_k$ is a cluster-specific parameter capturing the relationship between covariates and outcome.

%\subsection{Fundamental Assumptions}
% To extend this framework for causal effect-based supervision, i.e. $y_n\equiv\tau_n$, we assume that the following three fundamental assumptions hold: \emph{randomization}, i.e., every individual has the same non-zero probability of receiving either treatment or control; %formally, $\forall n, 0<p(a_n=1)<1$
% \citep{sävje2021randomization}, \emph{unconfoundedness}, i.e., the  treatment assignment is conditionally independent of the potential outcomes, given observed covariates, %i.e. $A\indep(Y^0,Y^1)\g X$, 
% and \emph{Stable Unit Treatment Value Assumption (SUTVA)}, i.e., the potential outcomes for any individual are unaffected by the treatment assignments of other individuals. 

    %\textbf{Positivity}: Every individual has a non-zero probability of receiving either treatment or control, conditional on covariates; formally, $\forall N, 0<p(A=1\g X=\bfx_n)<1$.

Consider we have observed data $\calD=\{\bfx_n,y_n^\obs,a_n\}_{n=1}^N$ where $a$ denotes the binary treatment assignment. We denote the potential outcomes for treatment $a_n=0$ and $a_n=1$ as $y_n^0$ and $y_n^1$ respectively, where either $y_n^0$ or $y_n^1$ are observed in practice. 
Since we do not observe treatment effects $\tau^\obs_n = y^1_n - y^0_n$, we cannot directly use this as outcome in the supervised clustering framework.
Instead, we model $p(y^\obs,a,\bfx)$, and parameterize each cluster based on their covariates and treatment effect, i.e.,
\begin{align*}
p(y^\obs_n,a_n,\bfx_n) %&= \sum_{k=1}^K p(y_n\g\bfx_n, z_n=k)p(\bfx_n\g z_n=k)p(z_n=k)\\&
=\sum_{k=1}^K \pi_kp(\bfx_n \g \bftheta_k) p(a_n \g \bfx_n, \bfzeta) p(y^\obs_n\g a_n,\bfx_n, \bfphi, \bfbeta_k)
\end{align*}
where $p(a\g \bfx, \bfzeta)$ denotes the propensity score, and we assume it to be independent of clusters.

For simplicity, we assume \emph{randomization}, i.e., $p(a | \bfx) = p(a)$, meaning every individual has the same non-zero probability of receiving treatment \citep{sävje2021randomization} implying 
\begin{align*}
p(y^\obs_n,a_n,\bfx_n) %&= \sum_{k=1}^K p(y_n\g\bfx_n, z_n=k)p(\bfx_n\g z_n=k)p(z_n=k)\\&
&\propto\sum_{k=1}^K \pi_kp(\bfx_n \g \bftheta_k) p(y^\obs_n\g a_n,\bfx_n, \bfphi, \bfbeta_k).
\end{align*}
Randomization is a stricter condition that implies unconfoundedness, i.e., $(Y^0, Y^1) \indep A \mid X$. While we focus on randomized settings, the framework naturally extends to observational data by modelling the propensity score explicitly \citep{Li2023BayesCausal}.

\paragraph{Potential Outcomes}  Assuming \emph{Stable Unit Treatment Value Assumption (SUTVA)} (i.e., no interference and no hidden variation of treatment),
\[y_n^\obs = y(a_n) = a_n y_n^1 + (1-a_n)y_n^0,\]
and assuming the outcome to be continuous, we parameterize the potential outcomes as  %$\E[\tau_n\g\bfx_n,z_n=k]=\E[y^1_n-y^0_n\g\bfx_n]=\E[y_n^\obs\g a_n=1,\bfx_n]-\E[y_n^\obs\g a_n=0,\bfx_n]=\tau(\bfx_n;\bfbeta_k)$,
\begin{align*}
y^0_n &= \mu^0(\bfx_n;\bfphi) + \epsilon_0\\
y^1_n &= \mu^0(\bfx_n;\bfphi)+\tau(\bfx_n;\bfbeta_k) + \epsilon_1
\end{align*}
with $\epsilon_0\sim\normal(0,\sigma_0^2)$, and  $\epsilon_1\sim\normal(0,\sigma_1^2)$. Additionally, $(\epsilon_0,\epsilon_1)\indep A$ due to randomization. Therefore,  
\begin{align*}
y_n^\obs\g a_n,\bfx_n,\bfphi,\bfbeta_k \distas \normal(\mu^0(\bfx_n;\bfphi)+a_n&\tau(\bfx_n;\bfbeta_k),a_n\sigma_1^2+(1-a_n)\sigma_0^2).
\end{align*}
Here $\tau(\cdot;\bfbeta_k)$ captures the treatment effects at each cluster $k$ while the control outcome $\mu^0(\cdot;\bfphi)$ does not depend on the cluster.
%Finally, we also consider the \emph{Stable Unit Treatment Value Assumption (SUTVA)}, i.e., the potential outcomes for any individual are unaffected by the treatment assignments of other individuals. 

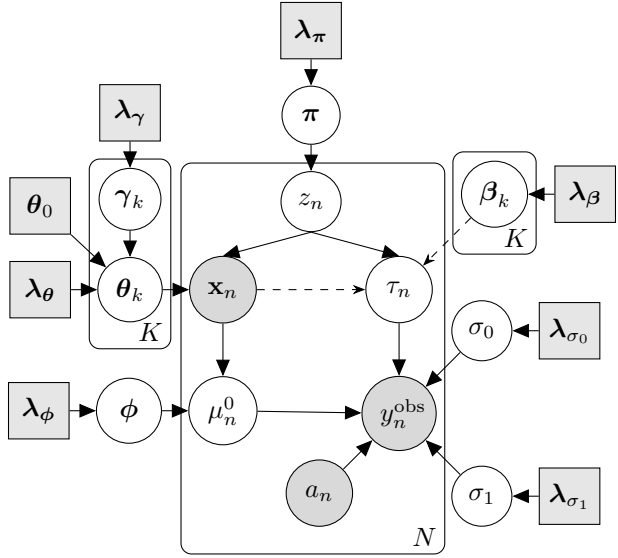
\begin{figure}[t]
    \centering
     \begin{tikzpicture}[node distance = 6mm]
    % Pis
    \node[latent] (pi) at (0, 0) {\(\bfpi\)};
    % Z
    \node[latent] (Z) [below=0.35cm of pi] {\(z_n\)};
    \node[latent, text width=0.5cm] (del_mu) [below right=0.8cm of Z] {\(\tau_n\)};%
    % Observations
    \node[observed] (X) [below left=0.8cm of Z] {\( \bfx_n \)};
    \node[observed] (y_obs) [below=0.7cm of del_mu] {\( y_n^\obs \)};
    \node[observed] (A) [below left=0.55cm of y_obs] {\( a_n\)};
    %\node[bayesnode] (y_0) [below= of X] {\( y_n^0 \)};
    %\begin{scope}[on background layer]
      %\fill[fill=gray!30] (y_0.90) arc [start angle=90, end angle=270, radius=0.45cm];
    %\end{scope}
    %\node[bayesnode] (y_1) [below = of del_mu] {\( y_n^1 \)};%
    %\begin{scope}[on background layer]
      %\fill[fill=gray!30] (y_1.90) arc [start angle=90, end angle=270, radius=0.45cm];
    %\end{scope}
    % latent variables / unobserved parameters
    \node[latent] (sigma0) [above right =0.65cm of y_obs] {\(\sigma_0\)};
    \node[latent] (sigma1) [below right=0.65cm of y_obs] {\( \sigma_1 \)};
    \node[latent] (del_coef) [above right=0.9cm of del_mu] {\( \bfbeta_k \)};%
    \node[param] (lambda_pi) [above=0.35cm of pi] {\( \bflambda_{\bfpi} \)};
    \node[param] (lambda_del_coef) [right=0.35cm of del_coef] {\( \bflambda_{\bfbeta} \)};
    \node[latent] (y0_mu) [below =0.7cm of X] {\( \mu_n^0 \)};
     \node[latent] (lambda_y0_mu_1) [left=0.35cm of y0_mu] {\( \bfphi \)};
      %\node[param] (lambda_y0_mu) [left=0.35cm of y0_mu] {\( m_0 \)};%below=0.4cm of y0_mu
    \node[param] (prior_lambda_y0_mu_1) [left=0.35cm of lambda_y0_mu_1] {\( \bflambda_\bfphi \)};
    \node[param] (lambda_sigma0) [right=0.35cm of sigma0] {\( \bflambda_{\sigma_0} \)};
    \node[param] (lambda_sigma1) [right=0.35cm of sigma1] {\( \bflambda_{\sigma_1} \)};
    \node[latent] (x_param_k) [left =0.35cm of X] {\( \bftheta_k \)};
    \node[latent] (gamma_k) [above=0.35cm of x_param_k] {\( \bfgamma_k \)};
    % known Parameters
    \node[param] (lambda_x_param) [left=0.35cm of x_param_k] {\( \bflambda_{\bftheta} \)};
    \node[param] (X_mean) [above=0.35cm of lambda_x_param] {\( \bftheta_0 \)};
    \node[param] (lambda_gamma) [above=0.35cm of gamma_k] {\( \bflambda_{\bfgamma} \)};
    
    % Edges
    \edge {lambda_pi} {pi};
    \edge {pi.south} {Z.north};
    \edge {Z.south} {X.north,del_mu.north};%
    \edge {del_mu.south} {y_obs.north};
    \edge {A} {y_obs};%
    \edge[dashededge] {X.east} {del_mu.west};%
    \edge {X} {y0_mu};%
    \edge {x_param_k}{X};
    \edge[dashededge] {del_coef}{del_mu};%
    \edge {y0_mu}{y_obs};
    \edge {sigma0}{y_obs};
    \edge {sigma1}{y_obs};
    \edge {lambda_x_param} {x_param_k};
    \edge {gamma_k} {x_param_k};
    \edge {X_mean} {x_param_k};
    \edge {lambda_gamma} {gamma_k};
    \edge {lambda_del_coef} {del_coef};
    %\edge {lambda_y0_mu} {y0_mu};
    \edge {lambda_y0_mu_1} {y0_mu};
    \edge {prior_lambda_y0_mu_1} {lambda_y0_mu_1};
    \edge {lambda_sigma0} {sigma0};
    \edge {lambda_sigma1} {sigma1};
    % Fit background boxes for clarity
    \plate [inner sep=.1cm,yshift=.01cm] {plate1}{(x_param_k)(gamma_k)} {\(K\)};
    \plate [inner sep=.1cm,yshift=.01cm] {plate2}{(del_coef)}{\(K\)};%xshift=.02cm,yshift=.2cm
    \plate [inner sep=.1cm] {plate3}{(Z)(X)(y_obs)(del_mu)(A)} {\(N\)};
\end{tikzpicture}
    \caption{Plate Diagram of {\basiccs} with feature selection. %: $\mu_n^0$ has a Gaussian process prior. %The potential outcomes for each group ($y^0_n$ and $y^1_n$) are only partially observed. 
    %$\tau_n$ is the individual treatment effect. 
    %The hyperparameters %of priors or hyperprior 
    %are denoted by $\bflambda_\cdot$.
    }
    \label{fig:supervised causal clust diagram GP}
\end{figure}

\paragraph{Generative Process}   Thus, given $K$ clusters, we consider the following generative framework (Figure \ref{fig:supervised causal clust diagram GP}),
\begin{align*}
    z_n &\distas\categorical(\bfpi)\\
    \bfx_n &\distas f(\bftheta_{z_n}) \\
    \mu^0_n &\quad=\quad\mu^0(\bfx_n; \bfphi) \\
    \tau_n  &\quad=\quad\tau(\bfx_n; \bfbeta_{z_n}) \\
    y_n^\obs&\distas \normal(\mu_n^0+a_n\tau_n,a_n\sigma_1^2+(1-a_n)\sigma_0^2).
    %y^0_n \g \bfx_n &\distas \normal(\mu^0_n, \sigma_0^2)\\
    %\tau_n \g z_n=k &\quad=\quad \tau(\bfx_n;\bfbeta_k)\\
    %y^1_n \g \bfx_n, z_n=k & \distas \normal(\mu^0_n+\tau_n,\sigma_1^2)
\end{align*} %}
 where $f(\bftheta_k)$ represents a density function of mixed-type covariates, where we assume a normal distribution, i.e., $\normal(\theta^\mu_{dk}, \theta^\sigma_{dk})$ for $d$-th continuous variable, and a Bernoulli (or multinomial for categorical) distribution, i.e., $\bernoulli(\theta^p_{dk})$ for $d$-th binary variable. %$\mu^0(\bfx_n)$ models the control outcome while $\tau(\bfx_n;\bfbeta_k)$ models the treatment effect over each cluster separately. %The potential outcomes are, albeit,  partially observed, i.e., for each individual, only $y_n^0$ or $y_n^1$ is observed but not both. 
We do not explicitly model $p(a)$ due to randomization. % with $\sigma^2_0$ and $\sigma^2_1$ being noise variances for $\epsilon_0$ and $\epsilon_1$ respectively. 
 
%The $k$-th subgroup-specific ATE under the above assumptions is then
%\begin{align*}
%\tau_k=\E[\tau(\bfx;\bfbeta_k)\g \bfx\in\calC_k].
%\end{align*}

\paragraph{Adaptation to Binary Outcome}
%The potential outcomes can also be assumed to be binary, e.g., ``death", i.e. $y^1, y^0\in\{0,1\}$. 

The outcome can also be assumed to be binary, e.g., ``death'', i.e., $y^\obs\in\{0,1\}$. In this context, we modify the output distributions as follows,
%{\small
% \begin{align*}
%     y^0_n \g \bfx_n &\distas \bernoulli(\inv{\logit}(\mu^0_n))\\
%     y^1_n \g \bfx_n, z_n=k & \distas \bernoulli(\inv{\logit}(\mu^0_n+\tau_n))
% \end{align*} %}
\begin{align*}
    %y^0_n &\distas \bernoulli(\inv{\logit}(\mu^0_n))\\
    %y^1_n & \distas \bernoulli(\inv{\logit}(\mu^0_n+\tau_n)) \\
    y^\obs_n & \distas \bernoulli(\inv{\logit}(\mu^0_n+ a_n\tau_n))
\end{align*}
where the individual treatment effect is given as $\tau_n=\log p^1_n/(1-p^1_n)-\log p^0_n/(1-p^0_n)=\log\{p^1_n/(1-p^1_n)\}/\{p^0_n/(1-p^0_n)\},$
i.e., the log odds ratio (log OR) if we define the individual probability of death under control and treated groups to be $p^0_n$ and $p^1_n$ respectively.
 
 % are two functions for relationships we mentioned above, representing individual mean control outcome and mean difference (i.e. treatment effect), which can be fitted by user-defined models. We can use a linear function or Gaussian process for a nonlinear function for $\mu^0(\bfx_n)$. We specify $\tau(\bfx_n,\bfbeta_k)$ as either a constant or linear function in each cluster in our model implementation. 
 
 % have the latent cluster variable $z_n\in\{1,\dots,K\}$ for the cluster assignment of individual $n$. We denote $\{\bftheta_k,\bfbeta_k\}_{k=1}^K$ as parameters of covariates distribution ($\bftheta_k$) and coefficients of the relationship between covariates and treatment effects ($\bfbeta_k$) in each cluster. $\mu^0(\bfx_n)$ is the function of the relationship between covariates and control outcomes. $\sigma_0$ and $\sigma_1$ are noise standard deviations for control and treated outcomes respectively. Therefore, the expressions of this supervised causal mixture model for HTE are:

\paragraph{Gaussian Process Prior on Control Outcome}
We use a Gaussian process (GP) to fit the nonlinear relationship $\mu^0(\bfx;\bfphi)$. A Gaussian process can be interpreted as providing a probability distribution over functions i.e. $\mu^0(\bfx)\sim\GP(m_0(\bfx),\kappa_0(\bfx,\bfx^\prime))$. For $N$ noise-free potential outcomes $\{\mu^0_n\}_{n=1}^N$ paired with $\{\bfx_n\}_{n=1}^N$ input covariates, it is a multivariate Gaussian on these finite number of outcomes $\bfmu^0\in\reals^N$ conditioned on their covariates:
$
\bfmu^0\g\bfX \sim \normal_N(\bfm_0,\bfK_0(\bfX))
$
where $\bfm_0=\bfzero$ is an $N$-vector of unknown function values and $\bfK_0(\bfX)$ is an $N\times N$ covariance matrix. We use the squared exponential kernel with noise for the covariance function, %that is:
%\begin{align*}
%    K(\bfx_i,\bfx_j\g\alpha,\rho,\sigma_0)&=\alpha^2\exp\left(\frac{-\norm{\bfx_i-\bfx_j}^2}{2\rho^2}\right)%+\I(i=j)\sigma_0^2
%\end{align*}
with the marginal standard deviation ($\alpha$), controlling the overall scale of function values and length-scale ($\rho$), determining how far apart inputs must be before their function values become uncorrelated. To capture varying influence across covariates, we adopt separate length scales for each dimension, following the automatic relevance determination (ARD) approach \citep{Neal1996ARD}. Thus, GP parameters $\bfphi=\{\alpha,\bfrho\}$. %A small $\rho$ closer to zero results in rapidly varying functions, while a large $\rho$ leads to smoother functions. $\sigma_0$ is the scale of the noise term. 
 Note that it is possible to use a linear kernel for GP to consider a simpler linear regression fit.

\paragraph{Choices for Treatment Effect} For better interpretability, we consider a constant treatment effect over each cluster $\tau(\bfx;\bfbeta_k) = \tau_k = \beta_k$. This homogeneity assumption is analogous to conventional risk stratification approaches in clinical trials, where sample sizes are small and constant treatment effects are commonly assumed across risk strata \citep{Kent2020PATH}. We can also consider a linear treatment effect \citep{Shahn2017LatentClass}, i.e., $\tau(\bfx;\bfbeta_k) = \bfbeta_k^\top\bfx$, but focus on the former in this paper. Intuitively, we jointly cluster samples based on their covariates and their treatment effect as shown in Figure \ref{fig:demo-simulated_data2D_byGMM&HTEclustX}\textbf{(b, right)}. The assumption of a normal distribution and a constant treatment effect facilitates interpretability and coherence by yielding clusters with spherical geometry and within each cluster, individuals share the same treatment-response characteristics.

\paragraph{Feature Selection}
Following the approach by \citet{Liverani2015FeatureSelection}, we employ the cluster-specific soft feature selection approach. Each mixture component $k$ has an associated vector $\bfgamma_k=(\gamma_{k,1},\dots,\gamma_{k,D})$, where $\gamma_{k,d}$ is a latent variable in $(0,1)$ that determines whether the $d$-th covariate is important to mixture component $k$. 
For the binary covariates, let $\theta^p_{d0}$ be the sample prevalence of binary covariate $d$ over the population, then we define the new composite cluster parameters as
$\bar{\theta}^p_{dk} \coloneq \gamma_{k,d}\theta^p_{dk}+(1-\gamma_{k,d})\theta^p_{d0}.$
This expression is substituted in place of $\theta^p_{dk}$ to provide the likelihood for the covariate model. %We assume that, given $r_d$, the $\gamma_{k,d}$ for $k=1,\dots,K$, are independent Bernoulli variables with $\gamma_{k,d}\sim\bernoulli(r_d)$. 
We further consider a weakly informative prior for $\gamma_{k,d}\sim\betarand(0.5,0.5)$.
%$$\gamma_{k,d} \distas \bfone_{(w_d=0)}\delta_0(\gamma_{k,d})+\bfone_{(w_d=1)}\betarand(0.5,0.5)$$where $w_d\sim\bernoulli(0.5)$ \citep{Liverani2015FeatureSelection}.
Similarly, for continuous covariates, we define $\theta^\mu_{d0}$ to be sample mean of continuous covariate $d$, and define new cluster parameters as
$\bar{\theta}^\mu_{dk} \coloneq \gamma_{k,d}\theta^\mu_{dk}+(1-\gamma_{k,d})\theta^\mu_{d0}.$
%The corresponding plate diagram for {\basiccs} is shown in Figure \ref{fig:supervised causal clust diagram GP}.

\paragraph{Implementation}
We implement {\basiccs} in \textsc{rstan} \citep{Rstan} using Automatic Differentiation Variational Inference (ADVI) \citep{Kucukelbir2017ADVI}. We use  non-informative priors for $\bfpi$ and $\bftheta_k$, half-normal priors with mean $0$ and small standard deviation ($0.01$) for the noise standard deviations $\sigma_0, \sigma_1$ to regularize them toward small values, and $\beta_k \sim \normal(0, 1)$ for the treatment effect by default (sensitivity analysis in Suppl.\ Tables~\ref{tab:sensitivity sd} and \ref{tab:sensitivity mean}). The GP hyperparameters $\bflambda_\bfphi\;(\bfphi=\{\alpha,\bfrho\})$ are set by first fitting a GP to the control outcome data via maximum likelihood estimation \citep{DiceKriging}, and cluster-specific parameters $\bftheta_k$ are initialized using a Gaussian Mixture Model ({\gmm}). Since Bayesian mixture models are known to suffer from posterior multimodality \citep{RamsésH2015Identifiability, stephensMultimodality2000, CarreiraModesGMM2003}, which can trap variational optimization in poor local optima, we run ADVI in parallel with diverse random initializations and select the solution with the highest evidence lower bound (ELBO).

\section{Simulation Results}\label{sec:sim res}

\paragraph{Metrics} We calculate the Adjusted Rand Index (\textsc{ARI}) to assess the clustering performance in the covariate space. Since different methods produce results in different forms, we report the range of Subgroup-specific Average Treatment Effects (SATEs) on the test data as a common metric to evaluate how well each method separates subgroups with distinct treatment effects. SATE is computed after assigning individuals to their respective clusters, and we expect that supervised causal clustering with treatment effect as supervision (e.g., {\basiccs}) should find clusters with more distinct treatment effects. Additionally, to assess the performance of ITE  estimation, we compute the Precision in Estimation of Heterogeneous Effects (\textsc{PEHE}) \citep{Hill2011BART}: $\textsc{PEHE}=\sqrt{\frac{1}{N}\sum_{i=n}^N(\tau_n-\E[y_n^1-y_n^0\g\bfx_n])^2}$. %For comparison with supervised learning methods, we assign cluster labels to individuals and treat the corresponding subgroup-specific treatment effect as $\tau_{Gk}$.

\paragraph{Null Treatment Effects ({\simNull})} We evaluate {\basiccs} under a setting where all individuals share a single cluster with zero treatment effect. Since {\basiccs} clusters jointly over covariates and treatment outcomes, it may still identify covariate-driven clusters when no treatment heterogeneity exists. As shown in Suppl.\ Table~\ref{tab:homo null TE}, when $K>1$, the model finds clusters with divergent estimated treatment effects $\tau_k$ on training data (e.g., range $[-0.69, 1.23]$ for $K=4$), but the corresponding test SATE ranges remain close to zero (e.g., $[-0.14, 0.03]$), indicating that the apparent heterogeneity does not generalise. Moreover, $K=1$ yields the lowest test \textsc{PEHE}, confirming the absence of heterogeneity. This highlights the importance of model selection: choosing the correct $K$ via validation avoids false positive discovery of spurious treatment effect heterogeneity.

\paragraph{Feature Selection ({\simFS})} To illustrate this, %the effectiveness of the feature selection for clustering, 
we simulated a two-dimensional dataset with four clusters but two different constant treatment effect groups as shown in Suppl.\ Figure \ref{fig:HTEclustXsim1100 feature imp visal}. In this case, only the first dimension or feature plays a significant role in clustering when considering treatment effects as the outcome, since the cluster means of the second feature are both close to the sample mean. The learned feature importance with {\basiccs} for each cluster is shown in Suppl.\ Figure \ref{fig:HTEclustXsim1100 feature imp}, where we see relatively larger weight for the first feature ($>0.5$) and small weight for the second feature ($<0.1$) with an \textsc{ARI} of $0.88$.

\begin{figure*}[!ht]
    \centering
    \includegraphics[width=\linewidth]{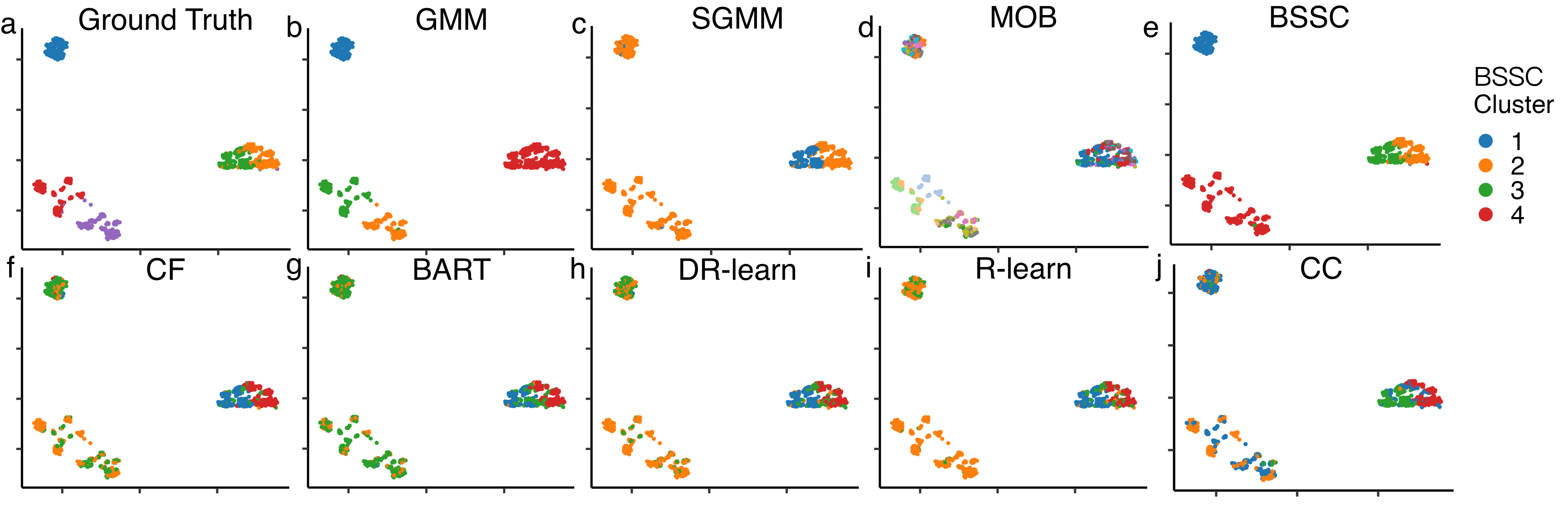}
    \caption{Visualization of UMAP Projection of the Simulated Dataset in {\simHTE}: Cluster assignments are given by \textbf{(a)} ground truth of constant treatment effects $(0.5,5,-5,0,0)$, \textbf{(b)} {\gmm}, \textbf{(c)} {\sgmm}, \textbf{(d)} {\mob}, \textbf{(e)} {\basiccs}, \textbf{(f)} {\cf}, \textbf{(g)} {\bart}, \textbf{(h)} \textsc{DR-learner}, \textbf{(i)} \textsc{R-learner} and \textbf{(j)} {\cc}.}
    \label{fig:HTEclustXsim1200 res}
\end{figure*}

\paragraph{Subgroup Discovery with HTE ({\simHTE})} We simulate a toy dataset consisting of $N=1200$ samples with $12$ covariates, including $6$ binary variables, to show the difference among various methods. Control outcomes are generated using a nonlinear function (detailed in Suppl.\ \ref{app:simHTE}), and the treated group comprises $50\%$ of the population. 
We manually construct $5$ clusters with specific properties, as visualized in Figure \ref{fig:HTEclustXsim1200 res}\textbf{(a)}: (1) Clusters 2 ($\tau_2=5$, orange) and 3 ($\tau_3=-5$, green) are close in covariate space but have opposite treatment effects; (2) Clusters 4 ($\tau_4=0$, red) and 5 ($\tau_5=0$, purple) are well-separated yet share the same null treatment effect; and (3) Cluster 1 ($\tau_1=0.5$, blue) is distinct from the others but exhibits a near-null treatment effect similar to Clusters 4 and 5. %Therefore, when considering the treatment effect, two clusters with the same null treatment effect may be merged. 
Our goal is to recover $4$ clusters with distinct treatment effects, as Clusters 4 and 5 share a null effect. Additionally, we generate a parallel dataset with identical settings but a smaller proportion of treated units ($20\%$) to evaluate sensitivity to treatment assignment imbalance. We split the data equally into training and test sets.

\begin{table}[t]%[!hbt]
    \centering
    \setlength{\tabcolsep}{4pt}
    \caption{Summary of performance for {\simHTE}.}\label{tab:synthetic dataset & clust res} 
    \begin{tabular}{lccc}\\\toprule
         Model (Treat Prop.) &  \textsc{ARI} & \textsc{SATE} & \textsc{PEHE}
         %Model &  \textsc{ARI} & Test Range of & Test\\
        %(Treat Prop.) & & SATE & \textsc{PEHE}
        \\\midrule
        {\gmm} (0.5) & \textbf{0.768}
        & [$-0.40$, $1.32$] & 3.13\\
        {\sgmm} (0.5) & 0.625 & [$-5.54$, $0.67$] & 2.65\\\midrule
        {\itt} (0.5) & 0.574 & [$-0.03$, $-0.03$] & 3.09\\
        {\mob} (0.5)& 0.437 & [$-4.73$, $2.38$] & 2.58\\\midrule
        {\basiccs} (0.2) & 0.715
        & [$-4.99$, $3.98$] & 1.49\\
        {\basiccs} (0.5) & 0.721
        & [$-5.13$, $3.99$] & \textbf{1.45}\\\midrule
        {\cf} (0.5) & 0.544 & [$-4.71$, $3.67$] & 1.98 \\
        {\bart} (0.5) & 0.381 & [$-4.46$, $4.61$] & 1.86 \\
        \textsc{DR-learner} (0.5)& 0.548 & [$-4.56$, $4.25$] & 2.14 \\
        \textsc{R-learner} (0.5)& 0.508 & [$-4.78$, $3.57$] & 1.72 \\
        {\findit} (0.5) & 0.126 & [$-0.33$, $0.31$] & 3.95\\
        {\owe} (0.5) & 0.333 & [$-3.91$, $2.22$] & 2.46\\\midrule
        {\cc} (0.5) & 0.524 & [$-3.72$, $3.76$] & 1.86\\
        \bottomrule
    \end{tabular}
\end{table}

Table \ref{tab:synthetic dataset & clust res} summarises the results from {\simHTE}, evaluating the performance of {\basiccs}, along with comparisons to (1) {\gmm} and (2) {\sgmm} with treated outcome as supervision only. Moreover, we also benchmark {\basiccs} against widely used supervised learning approaches: (3) {\itt}, (4) {\mob}, (5) {\cf}, (6) {\bart}, (7) \textsc{DR-learner}, (8) \textsc{R-learner}, (9) {\findit}, (10) {\owe}, and (11) {\cc}, evaluated on balanced simulated datasets with treatment proportion $0.5$. Cluster labels for effect modeling and causal clustering methods are obtained by applying {\gmm} to estimated treatment effects or potential outcomes, respectively. For each method, we identify four clusters. ITEs for clustering-based methods are approximated by SATEs.

We find that {\basiccs} attains a moderate \textsc{ARI} of $0.721$, compared to $0.768$ for {\gmm}. However, {\gmm} produces SATE estimates that fail to reflect the true heterogeneity (range $[-0.40, 1.32]$ versus the ground truth $[-5, 5]$), highlighting a fundamental tension between clustering fidelity in covariate space and treatment effect estimation.
In terms of SATE, {\basiccs} produces estimates that closely match the ground truth (range $[-5.13, 3.99]$), demonstrating superior fidelity in capturing subgroup-level heterogeneity. Among all methods, {\basiccs} achieves the lowest test \textsc{PEHE} ($1.45$), outperforming the next-best \textsc{R-learner} ($1.72$), {\bart} and {\cc} ($1.86$), and {\cf} ($1.98$). Notably, methods such as {\itt} and {\findit} yield near-degenerate SATE ranges, indicating a failure to capture heterogeneity in the test data.
Performance for {\basiccs} remains robust when the treated proportion is reduced to $0.2$, with only a marginal increase in \textsc{PEHE} ($1.49$) and a slight decline in \textsc{ARI} ($0.715$).

%We select representative models from each category (see Supplementary Table~\ref{tab:subgroup_tasks}) for UMAP visualization based on their performance: GMM (unsupervised), SGMM (supervised with outcome), {\basiccs} (ours), CF, BART, DR-learner, R-learner (effect modeling), and MOB (subgroup analysis).

To explore the difference among the various methods, as illustrated in Figure \ref{fig:HTEclustXsim1200 res}\textbf{(b)}, {\gmm} does not distinguish between Clusters 2 ($\tau_2=5$) and 3 ($\tau_3=-5$), which are close in covariate space but have opposite treatment effects. In contrast, {\basiccs} in Figure \ref{fig:HTEclustXsim1200 res}\textbf{(e)} separates them, while merging Clusters 4 and 5 that share the same null treatment effect.

In Figure \ref{fig:HTEclustXsim1200 res}\textbf{(c)}, {\sgmm} supervises clustering using only the treated outcome $\E[y^1] = \E[y^0] + \tau$ with noise. This limitation is best understood via Suppl.\ Figure~\ref{fig:HTEclustXsim1200 Y0 Y1}: projecting onto the $y^1$ axis (vertical), Cluster 3 (green, $\tau_3=-5$) occupies a distinctly low range ($y^1 \approx -6$ to $1$), making it easily separable. However, the remaining clusters overlap substantially along $y^1$: Cluster 2 (orange, $\tau_2=5$) has $y^1 \approx 4$--$12$, but because Clusters 2 and 3 share similar covariate distributions (and hence similar $y^0$), Cluster 2's high treated outcome $y^0 + 5$ overlaps with Cluster 4 (red, $\tau_4=0$), whose higher baseline $y^0$ yields $y^1 \approx 3$--$9$. Similarly, Clusters 1 and 5 overlap in $y^1$. Without access to $y^0$, {\sgmm} cannot disentangle baseline response from treatment effect, and thus merges Clusters 1, 2, 4, and 5. This is reflected in its SATE range $[-5.54, 0.67]$, recovering only the negative effect.

For effect modeling methods (Figure~\ref{fig:HTEclustXsim1200 res}\textbf{(f,g,h,i)}), many individuals are misclassified when clustering is based solely on estimated treatment effects. Subgroup analysis methods such as {\mob} (Figure~\ref{fig:HTEclustXsim1200 res}\textbf{(d)}) tend to over-partition the data, producing more clusters than the ground truth due to their one-variable-at-a-time splitting strategy, and fail to capture interactions across multiple covariates. Causal clustering (Figure~\ref{fig:HTEclustXsim1200 res}\textbf{(j)}) clusters on imputed potential outcome vectors $(y^0, y^1)$; while it captures some structure, it cannot distinguish clusters that share similar potential outcomes but differ in covariate space (e.g., Clusters 1, 4 and 5), as further illustrated in Suppl.\ Figure~\ref{fig:HTEclustXsim1200 Y0 Y1}.

\textbf{Prior Sensitivity of $\beta_k$} We use the same simulated dataset as {\simHTE}. As shown in Table~\ref{tab:sensitivity sd}, the model is robust to the choice of prior standard deviation: varying the SD from $0.5$ to $100$ yields stable \textsc{ARI} ($0.73$--$0.75$) and \textsc{PEHE} ($0.85$--$0.99$). Providing informative prior means closer to the true effects can slightly improve performance (Table~\ref{tab:sensitivity mean}), though the default non-informative prior performs well.

\begin{figure*}[!ht]
  \centering
    \includegraphics[width=\linewidth]{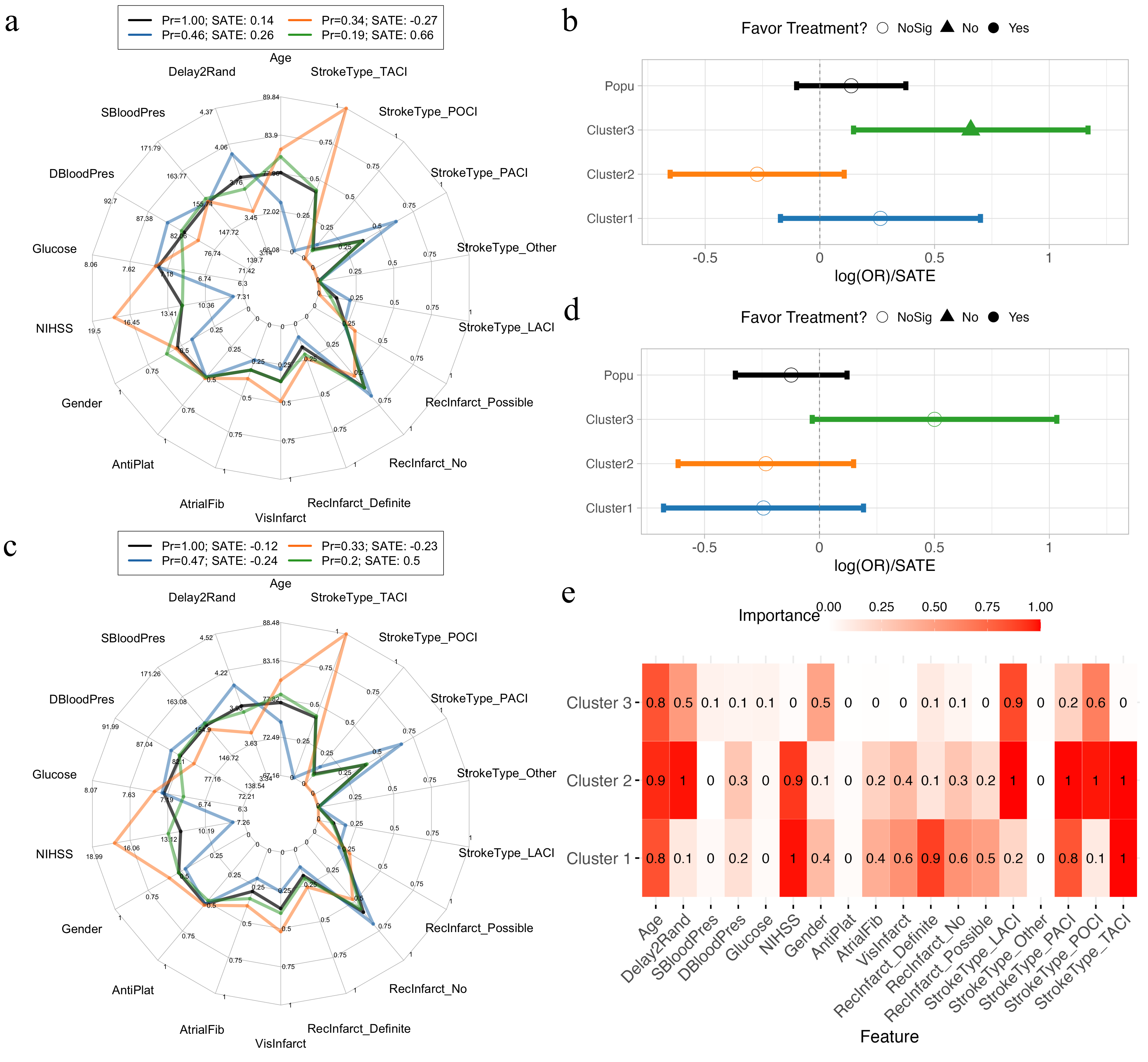}
  \caption{{\basiccs} on IST-3 Dataset: \textbf{(a)} The estimated cluster means and \textbf{(b)} corresponding ORs with $95\%$ CIs for training (in log scale), and \textbf{(c)} the empirical cluster means and \textbf{(d)} corresponding ORs with $95\%$ CIs for test set. Each axis representing a continuous covariate in the radar chart is transformed back to its original scale to enhance interpretability. \textbf{(e)} Cluster-specific feature importance.}
  \label{fig:IST3 causal super-Clust res}
\end{figure*}

\section{Real Application}

We assess the effectiveness of {\basiccs} on the Third International Stroke Trial (IST-3), evaluating the benefit of thrombolysis in patients with acute ischaemic stroke \citep{IST3Data2012}. Patients were either allocated to $\qty{0.9}{\mg/\kg}$ intravenous recombinant tissue plasminogen activator (rt-PA) or to control by placebo. The primary outcome used is the Oxford Handicap Score (OHS) at 6 months, which indicates if the patient is alive and independent, ranging from $0$ to $6$ with $6$ representing death. We selected $12$ features as covariates for clustering involving $6$ continuous variables, based on findings from \citet{IST3Data2012}, as detailed in Suppl.\ Table \ref{tab:IST-3 list of covariates}. After dropping missing values in features, we have a total of $2737$ patients. The proportion of observed OHS for control and treated groups is shown in Suppl.\ Figure \ref{fig:IST3_norm-Y distribution}. We standardised continuous covariates and performed one-hot encoding for categorical covariates having categories greater than $2$.

We split the dataset evenly into training and testing sets to ensure comparable sample sizes. To improve the modelling of control outcomes in the training set, we converted the outcome to a binary variable indicating ``death'', corresponding to OHS$=6$ in Figure \ref{fig:IST3_norm-Y distribution}, which has the proportion of around $26\%$. We assume a constant treatment effect within each cluster, allowing it to be interpreted as log odds ratio (log OR) for binary outcomes. Model performance is evaluated using two metrics on the test data: the accuracy of predicted control outcomes and the policy risk \citep{Shalit2017ITE-ICML}, which measures the expected outcome under a treatment policy $\textsc{Pol}(\bfx)$ that assigns each individual to the treatment arm with the better estimated effect. A lower policy risk indicates that the estimated treatment effects lead to better treatment decisions on held-out data: 
$R_{\textsc{Pol}}= 1-(\E[y^1\g\textsc{Pol}(\bfx)=1,a=1]p(\textsc{Pol}(\bfx)=1)+\E[y^0\g\textsc{Pol}(\bfx)=0,a=0]p(\textsc{Pol}(\bfx)=0))$.
%where $\textsc{Pol}(\bfx)$ is a treatment assignment rule that maps covariates to either treated or control, guiding individualized treatment decisions.

%\subsection{Clustering Results}
We identified three clusters that exhibit relatively high accuracy in predicting control outcomes and low policy risk, as reported in Suppl.\ Table \ref{tab:IST-3 select the number of clusters}. Since cluster assignments during testing are based solely on covariate information, it is important to assess the consistency between training and test results. Notably, when $K=3$, the model yields consistent cluster means and odds ratios (ORs) across both sets, as illustrated in Figure \ref{fig:IST3 causal super-Clust res}.

We visualise the estimated cluster means given by the training dataset compared with the empirical cluster means obtained by cluster assignments for the test dataset in Figure \ref{fig:IST3 causal super-Clust res}\textbf{(a)} and \textbf{(c)}. This helps with examining the differences and stability between learning and deploying this model. In addition, we also calculate the empirical ORs (SATE) with $95\%$ confidence intervals (CIs) for each cluster to evaluate the tendency of each cluster towards the treatment in Figure \ref{fig:IST3 causal super-Clust res}\textbf{(b)} and \textbf{(d)}.

%Based on the clustering results and detailed statistical results in Appendix \ref{app:real res},
The three identified clusters exhibit distinct prognostic profiles:
%\begin{enumerate}
    %\item 
    \textbf{Cluster 1} comprises patients with characteristics associated with a more favorable prognosis: younger age, lower NIH Stroke Scale (NIHSS) scores, a higher proportion with no visible ischaemia on CT scans, and predominantly milder stroke syndromes (partial anterior circulation and lacunar). This group has the lowest control arm mortality at $13.3\%$.
    %\item 
    Compared with the whole trial population, \textbf{Cluster 2} represents individuals with more severe strokes and additional poor prognostic indicators. These patients are older, have higher NIHSS scores, and almost exclusively exhibit total anterior circulation infarcts, the most severe stroke subtype. This cluster demonstrates the highest control arm mortality at $47.6\%$.
    %\item 
    \textbf{Cluster 3} largely reflects the overall trial population, but with a few distinctions: patients are moderately older, have lower blood glucose levels, and a higher proportion show definite ischaemia on CT scans, suggesting delayed presentation or diagnosis. The control arm mortality in this group is $22.8\%$, slightly lower than the overall trial population control group ($25.9\%$). 
%\end{enumerate}

Regarding feature importance, age and absence of lacunar stroke syndrome are most influential in Cluster 3. For Cluster 2, key features include age, delay to randomisation, NIHSS score, and stroke syndrome type. In Cluster 1, the most important variables are age, NIHSS, evidence of infarction on CT, and stroke subtype. 
These findings align with subgroup effects reported in the original trial by \citet{IST3Data2012}, where HTE were identified for age, NIHSS, and predicted probability of poor 6-month outcome (driven by age and NIHSS). The pronounced HTE in Cluster 3 may reflect its higher prevalence of older individuals and elevated NIHSS scores.

Suppl.\ Figure \ref{fig:IST3 GMM res} shows the clustering results by the unsupervised clustering approach using {\gmm} for comparison. We observe that when specifying $K=3$, the {\gmm} finds three clusters with log ORs very close to $0$ i.e. range of SATE $[-0.01,0.14]$. Although {\basiccs} provides improvement in terms of identifying different treatment effect tendencies i.e. range of SATE $[-0.27,0.66]$, the improvement is not statistically significant in the test dataset. Nevertheless, as shown in the figures in Suppl.\ \ref{app:real res}, supervised learning models often produce unstable subgroup structures, with log-odds ratio (SATE) estimates that vary considerably between training and test sets. The subgroup analysis method {\mob} identified four clusters using only the splitting covariates ``NIHSS" and ``age", while failing to account for clinically important covariates such as stroke subtypes, which were captured by {\basiccs}.
%This is because of a relatively low control outcome fitting given the covariates, which is typically greater than $0.85$ in simulations.

\section{Discussion}
% Comparison to Existing Methods
We presented a Bayesian supervised causal clustering framework ({\basiccs}) that simultaneously models covariates and treatment responses. %Unlike unsupervised clustering methods that ignore outcome information, or supervised learning models that lack subgroup structure, {\basiccs}  discovers interpretable clusters with both covariate similarity and consistent treatment effects. 
We apply {\basiccs} in the RCT population to find meaningful subgroups aligned to heterogeneous treatment effects. This has several advantages over existing methods.
While traditional subgroup analyses identify subgroups based on splitting covariate space or regressing with outcome labels, they do not directly find subgroups based on covariate similarities, resulting in clusters that do not capture interactions between covariates or overlapping clusters. Traditional clustering approaches, on the other hand, are capable of identifying similar covariate subgroups but may not align with treatment response heterogeneity. Causal clustering approaches directly cluster based on causal effects; however, the clustering structure might be less operationalizable. The supervised clustering methods utilize non-causal outcomes without considering causal structures. In contrast,  {\basiccs} explicitly incorporates treatment effects as outcome information, enabling principled identification of interpretable, operationalizable, and consistent subpopulations under treatment effect heterogeneity.

% Model Insights and Performance
Empirical evaluations on synthetic datasets demonstrate the framework's flexibility in handling mixed-type covariates with feature selections and nonlinear outcome models, as well as avoiding spurious discovery (discussed in {\simNull}).  Importantly, the model recovers meaningful cluster structures even when traditional clustering and supervised clustering methods fail. For example, it distinguishes covariate-similar subgroups with opposing treatment effects and merges covariate-distant groups with shared treatment responses, capabilities that standard {\gmm} and {\sgmm} baselines lack. In real-world clinical data from the IST-3 stroke trial, our {\basiccs} model uncovered three clinically meaningful clusters differentiated by age, stroke severity, and imaging biomarkers. These subgroups showed distinct mortality patterns and treatment responses, aligning with known clinical findings and supporting the model's utility in health applications.

% Limitations and Future Work
{\basiccs} can be extended in various manners.
Our current approach assumes randomized treatment assignment, and the approach can be extended to observational data settings. The current formulation considers a fully supervised set-up, and it can be extended  {\basiccs} to a semi-supervised clustering variation by disregarding the outcome information of unlabelled samples during training. We consider binary treatments with continuous or discrete outcomes, and the approach can be extended to consider multiple treatment arms and other outcome modalities, e.g., time-varying, time-to-event, etc. Finally, the method is currently implemented in ADVI, and can be extended to handle larger datasets with more tailored inference methods.

%\paragraph{Remarks on Inference and Theoretical Properties}
%Exact inference in this joint framework involving mixture models and Gaussian processes is analytically intractable.
%While ADVI provides a scalable approximation by minimizing the Kullback-Leibler (KL) divergence between the variational family and the true posterior, it is known to underestimate posterior variance \citep{blei2017variational}. Furthermore, unlike MCMC, ADVI does not provide asymptotic guarantees of converging to the exact posterior distribution.
%However, our primary focus is on identifying latent cluster structures and estimating mean treatment effects, tasks for which variational approximations have shown strong empirical performance \citep{Kucukelbir2017ADVI}.
%Theoretically, identifiability in mixture models is a known challenge \citep{stephensMultimodality2000};
%however, in our supervised setting, the additional signal from the treatment outcome constrains the posterior, helping to separate clusters that might overlap in covariate space but differ in treatment response.
%Given the complexity of deriving non-asymptotic bounds for this joint model, we rely on extensive empirical validation on simulated and real-world data to demonstrate the robustness and utility of {\basiccs}.

%\section*{Acknowledgments}
\bibliography{references}

\newpage

\appendix

\begin{center}
    \LARGE\textbf{Bayesian Supervised Causal Clustering\\(Supplementary Material)}
\end{center}
\vspace{1em}

\section{Additional Model Information}\label{app:info}

\begin{table}[htbp]
\centering
\small
\caption{Summary of Related Works}
\label{tab:subgroup_tasks}
\begin{tabular}{@{}lllll@{}}
\toprule
\textbf{Group based on} & \textbf{Task} & \textbf{Example} & \textbf{Covariate Split} & \textbf{Supervision} \\
\midrule
\multirow{4}{*}{\shortstack[l]{\textbf{Clustering}\\\textbf{covariates}}}
 & Unsupervised & {\gmm}, {\lca} & Multivariate & None \\
 & Supervised & BayesProfile, {\sgmm} & Multivariate & Outcome \\
 & Supervised & {\itt}, {\mob} & One-at-a-time & Treatment effect \\
 & Supervised & {\basiccs} & Multivariate & Treatment effect \\
\midrule
\shortstack[l]{\textbf{Clustering}\\\textbf{treatment effects}}
 & Supervised & \shortstack[l]{Meta-learners, {\cf},\\{\bart}, {\findit}, {\owe}} & None & Treatment effect \\
\midrule
\shortstack[l]{\textbf{Clustering}\\\textbf{potential outcomes}}
 & Supervised & Causal Clustering & None & Potential outcomes \\
\bottomrule
\end{tabular}
\end{table}

In our model, we assume: (1) multiple covariates are independent and continuous covariates are distributed by Gaussian while binary covariates are distributed by Bernoulli. (2) one binary treatment variable i.e. control or treated; (3) scalar continuous outcome variables for control and treated groups for supervision, although it is easy for our model to extend to binary outcome variables by applying a sigmoid function (binary outcome variable is used for the real dataset).
For example in tabular data, it might be Table \ref{tab:example tabular data}.
\begin{table}[!ht]
    \centering
    \caption{Example Tabular Data for HTE}
    \label{tab:example tabular data}
    \setlength{\tabcolsep}{1mm}
    \begin{tabular}{cccccc}\toprule
        ID & Gender  & Age  & Treated &  Control & Treated\\
        & & & & Outcome & Outcome \\\midrule
 1& F & 61 & Yes & NA & 2\\
 2& M & 35 & No & 1 & NA\\
 3& F & 46 & Yes & NA & 1\\\bottomrule
    \end{tabular}
\end{table}

\section{Additional Simulation Results}\label{app:sim res}

\subsection{Sanity Check} 
We simulate our test datasets for the simplest linear (\textsc{L}) or constant (\textsc{C}) cases of control outcome functions and treatment effects for a sanity check. The number of covariates used is $9$, including $3$ binary variables. The ground truth number of clusters for all four datasets is $5$. We calculate the Adjusted Rand Index (\textsc{ARI}) to assess the clustering accuracy as shown in Table \ref{tab:synthetic dataset & res sanity check}. The average relative $L_2$ error is computed by taking means of relative $L_2$ errors for $\{\{\bftheta_k,\bfbeta_k\}_{k=1}^K,\bfpi,\bfmu^0\}$, reflecting the estimation accuracy of parameters.

\begin{table}[!hbt]
    \centering
    \caption{A Summary of Synthetic Datasets and Model Performance for Sanity Check of {\basiccs}: The data case of \textsc{L-C} (i.e. linear-constant) means the data is generated from a linear control outcome function and constant treatment effects for each cluster.}
    \label{tab:synthetic dataset & res sanity check}
    \begin{tabular}{cccc}\\\toprule
         Size&  Data Case  & Avg. Rel. 
         & \textsc{ARI}\\
          & (Treated Prop.)  & $L_2$ Error &
          \\\midrule
         $N=720$ & \textsc{L-C} (0.5) & 0.04 
         & 1\\
        $N=740$  & \textsc{L-C} (0.2) & 0.033 
        & 0.996 \\
        $N=760$  & \textsc{L-L} (0.5) & 0.072 
        & 1 \\
        $N=780$  & \textsc{L-L} (0.2) & 0.054 
        & 0.996\\
        \bottomrule
    \end{tabular}
\end{table}

\begin{figure}[!ht]
    \centering
    \begin{subfigure}[b]{0.47\linewidth}
        \centering
        \includegraphics[width=\linewidth]{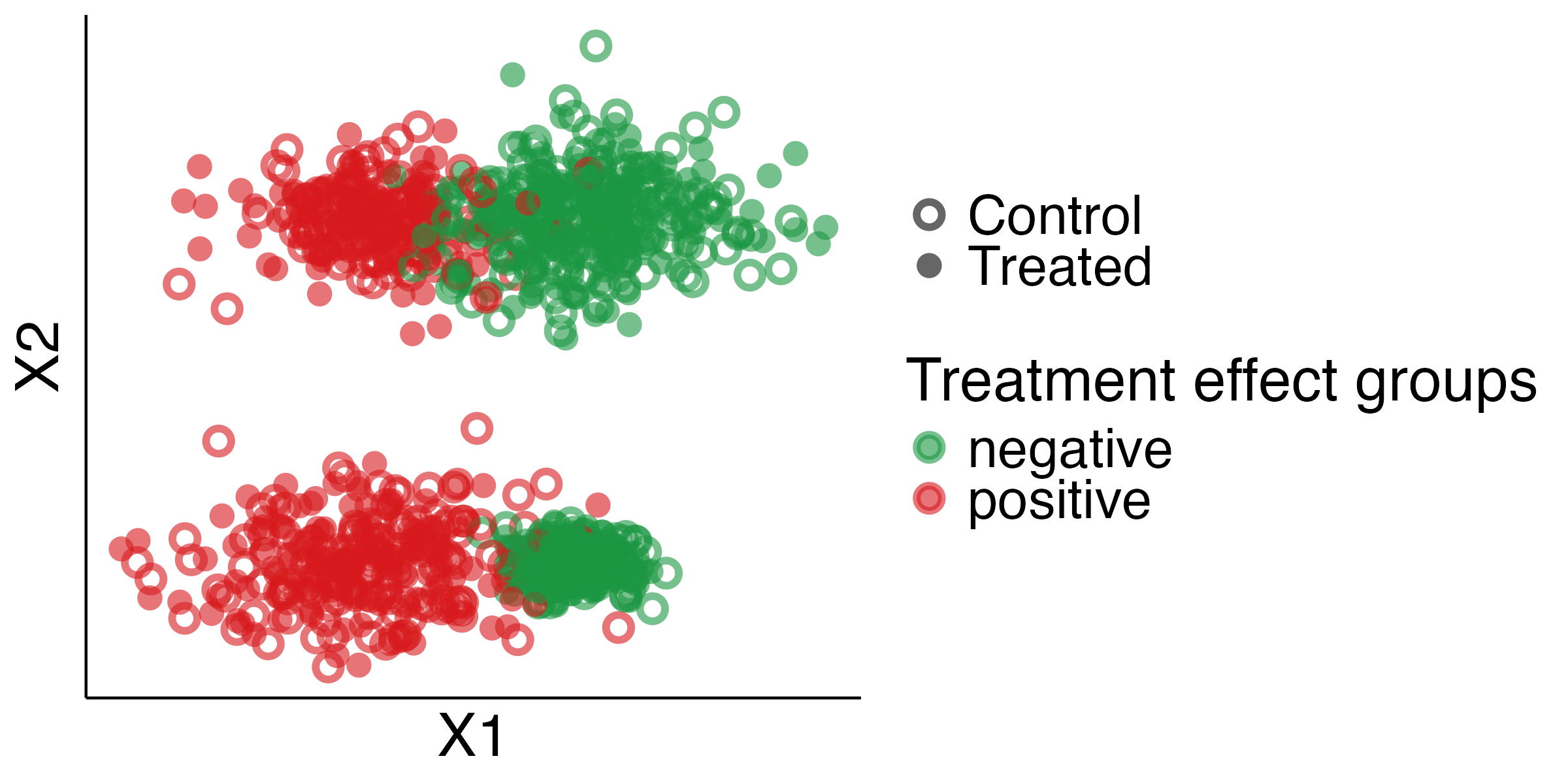}
        \caption{Simulated dataset}
        \label{fig:HTEclustXsim1100 feature imp visal}
    \end{subfigure}
    \hfill
    \begin{subfigure}[b]{0.45\linewidth}
        \centering
        \includegraphics[width=0.85\linewidth]{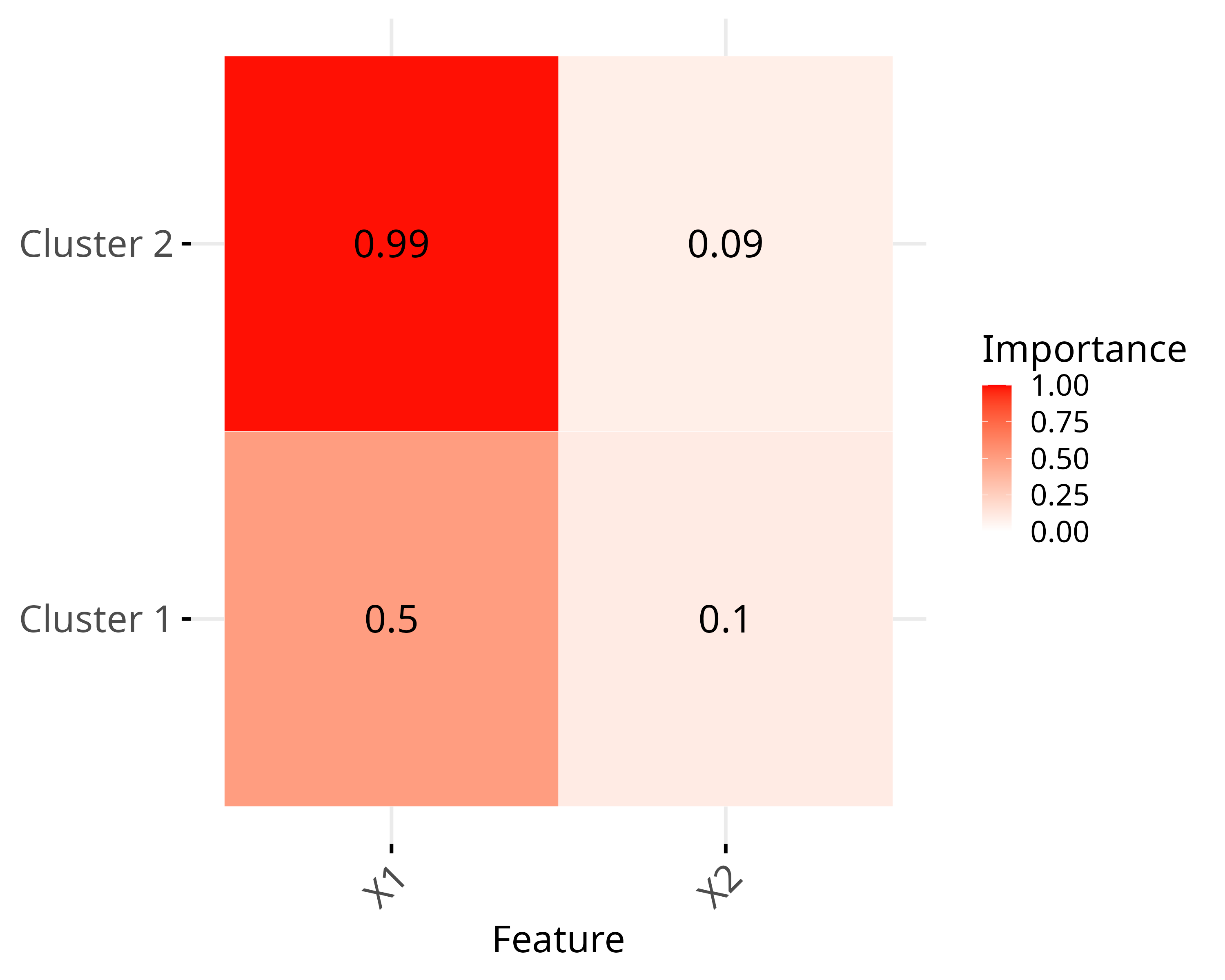}
        \caption{Feature importance}
        \label{fig:HTEclustXsim1100 feature imp}
    \end{subfigure}
    \caption{{\simFS}: \textbf{(a)} Simulated dataset and \textbf{(b)} learned cluster-specific feature importance by {\basiccs}.}
    \label{fig:HTEclustXsim1100 feature imp combined}
\end{figure}

Our model demonstrates strong performance across all four datasets, with small average relative $L_2$ error ($<0.08$) and \textsc{ARI}s all close to 1, which assumes a linear control outcome function, particularly when a linear kernel is used for the Gaussian process prior. Additionally, when adopting an exponentiated quadratic (RBF) kernel with carefully tuned hyperparameters, the model achieves near-perfect clustering and treatment effect estimation accuracy. For the demonstration of feature selection technique, Figure \ref{fig:HTEclustXsim1100 feature imp} shows the learned cluster-specific feature importance.

\subsection{HTE Comparison to Supervised Learning Methods}\label{app:simHTE}
The {\simHTE} dataset consists of $N=1200$ individuals with $D=12$ covariates ($6$ continuous, $6$ binary), drawn from $5$ clusters with mixing proportions $\bfpi \approx (0.19, 0.21, 0.17, 0.21, 0.22)$. Continuous covariates are sampled from cluster-specific multivariate Gaussians; binary covariates are drawn as independent Bernoulli variables. The control outcome follows a nonlinear function of the covariates:

{\small
\begin{align*}
\mu^0(\bfx) = &\;\sin(\pi x_1 x_2) + 0.2(x_3 - 0.5)^2 + \frac{\exp(x_4)}{1+\exp(x_5)} + x_6^2 + 0.3 x_7 + \log(1 + x_8 x_9) + 2(x_{10} - 0.5)^2 + x_{11} x_{12}
\end{align*}
}
with additive Gaussian noise ($\sigma=1$) for both $y^0$ and $y^1$. Constant treatment effects are assigned per cluster: $\bftau = (0.5, 5, -5, 0, 0)$, and the treated proportion is $0.5$ (with a parallel dataset at $0.2$).

\begin{figure}[!ht]
    \centering
    \includegraphics[width=0.6\linewidth]{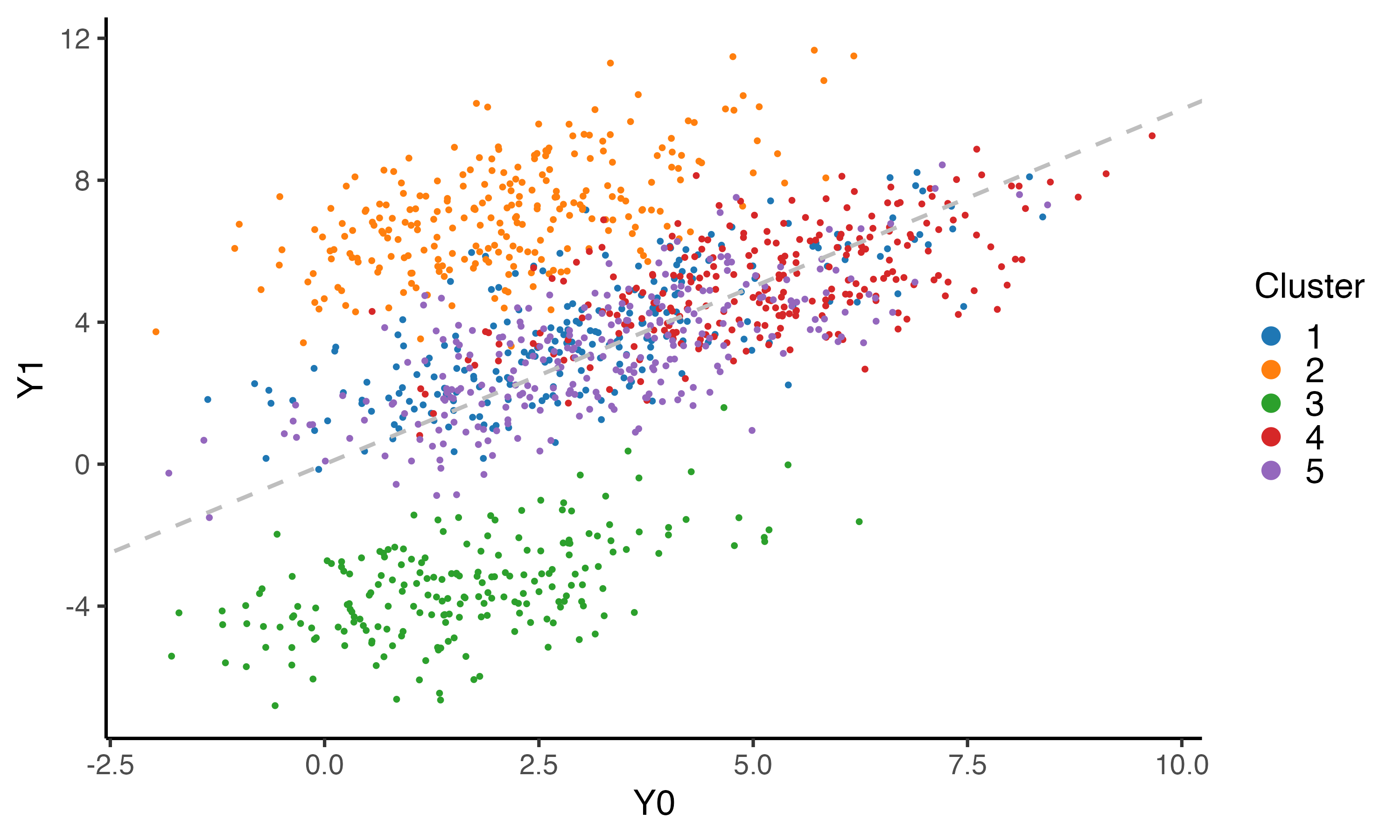}
    \caption{True potential outcomes $(y^0, y^1)$ for {\simHTE}, colored by ground-truth cluster membership. The dashed line indicates $y^1 = y^0$ (zero treatment effect).}
    \label{fig:HTEclustXsim1200 Y0 Y1}
\end{figure}

Figure~\ref{fig:HTEclustXsim1200 Y0 Y1} displays the true potential outcomes $(y^0, y^1)$ for each individual, colored by ground-truth cluster membership. The dashed line represents the $y^1 = y^0$ diagonal (i.e., zero treatment effect). Clusters 2 (orange, $\tau=5$) and 3 (green, $\tau=-5$) are clearly separated above and below the diagonal, respectively, reflecting their strong positive and negative treatment effects. In contrast, Clusters 1 (blue, $\tau=0.5$), 4 (red, $\tau=0$), and 5 (purple, $\tau=0$) lie close to the diagonal, consistent with their near-null effects. This visualization highlights the challenge faced by causal clustering methods that rely solely on potential outcome vectors: Clusters 4 and 5 overlap substantially in $(y^0, y^1)$ space despite being well-separated in covariate space, underscoring the advantage of {\basiccs} in leveraging covariate information alongside treatment effect supervision.

\begin{figure*}[!ht]
    \centering
    \begin{subfigure}[b]{0.32\linewidth}
        \centering
        \includegraphics[width=\linewidth]{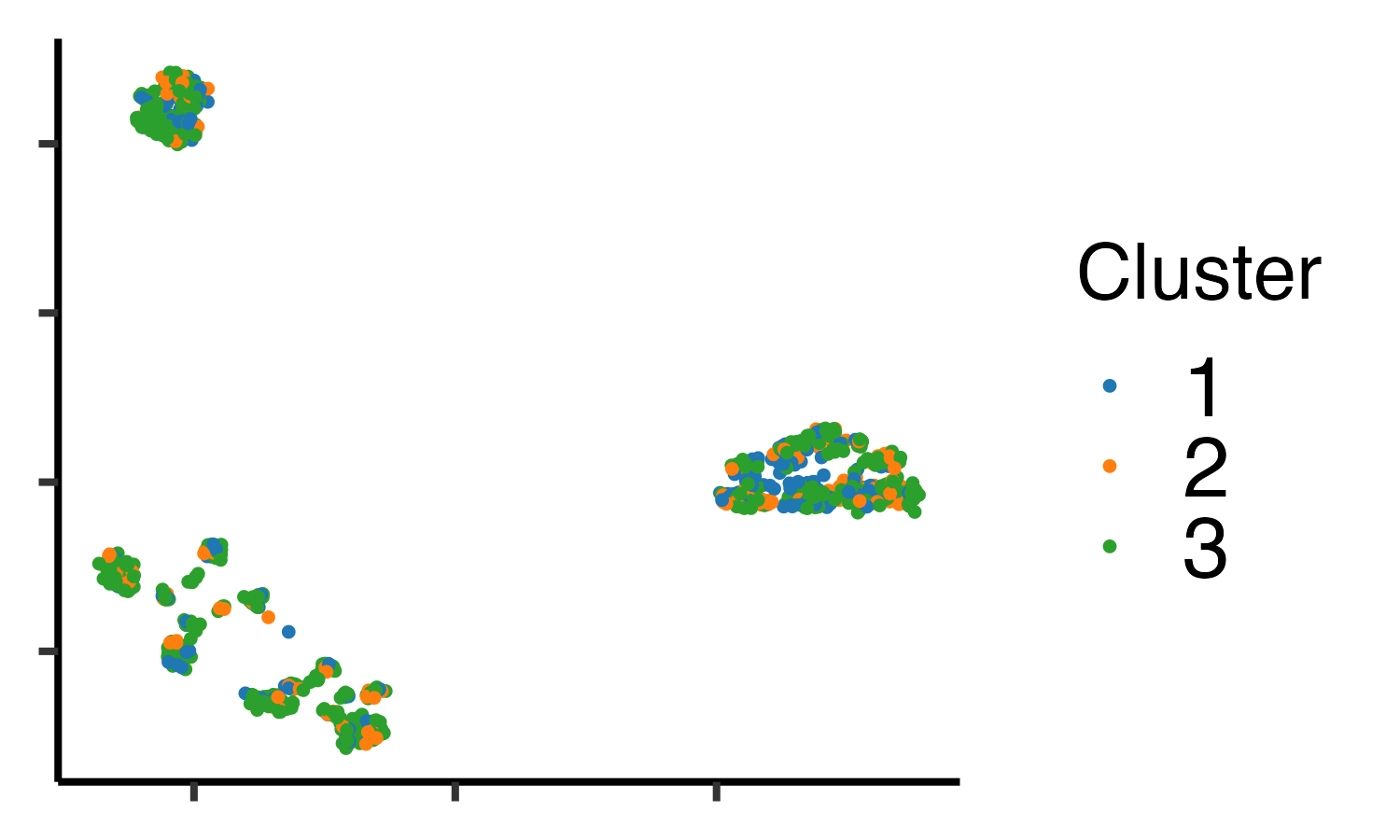}
        \caption{{\itt}}
        \label{fig:HTEclustXsim1200 UMAP IT}
    \end{subfigure}
    \hfill
    \begin{subfigure}[b]{0.32\linewidth}
        \centering
        \includegraphics[width=\linewidth]{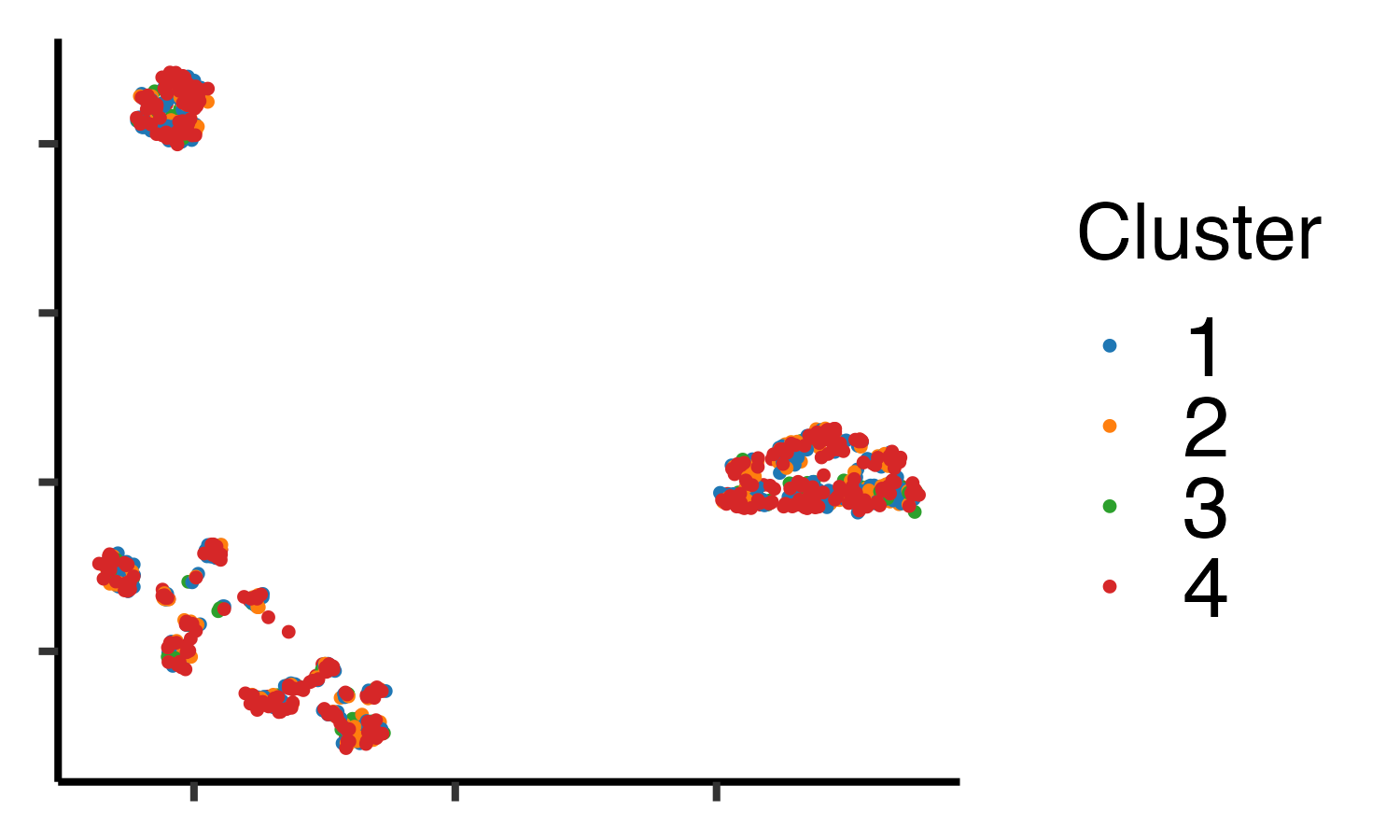}
        \caption{{\findit}}
        \label{fig:HTEclustXsim1200 UMAP FI}
    \end{subfigure}
    \hfill
    \begin{subfigure}[b]{0.32\linewidth}
        \centering
        \includegraphics[width=\linewidth]{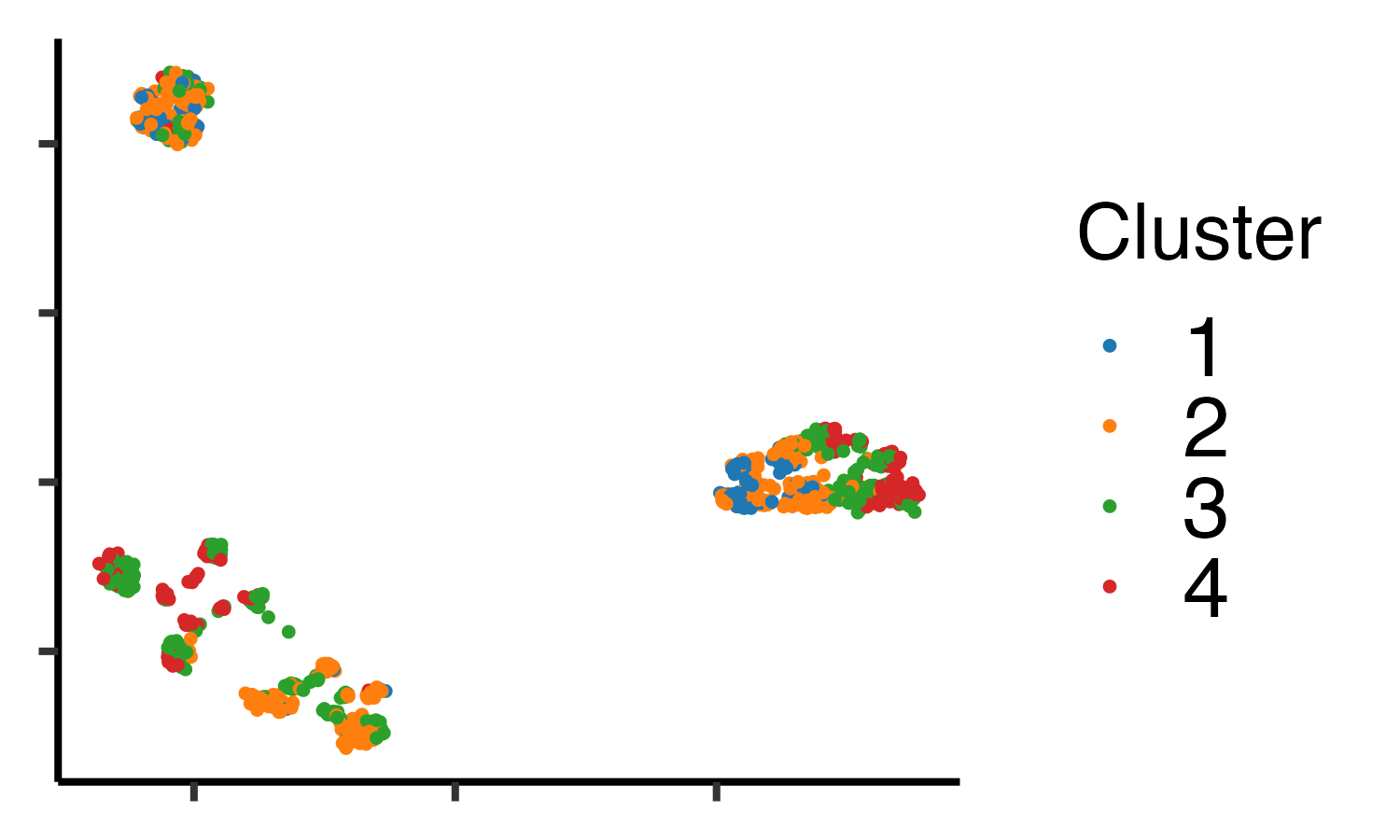}
        \caption{{\owe}}
        \label{fig:HTEclustXsim1200 UMAP OWE}
    \end{subfigure}
    \caption{UMAP projections of covariate space for {\simHTE}, colored by cluster assignments from \textbf{(a)} {\itt}, \textbf{(b)} {\findit}, and \textbf{(c)} {\owe}.}
    \label{fig:HTEclustXsim1200 UMAP extra}
\end{figure*}

We show the comparison results of our model to common supervised learning methods for the heterogeneous treatment effects listed in the main text. Since supervised learning methods, except subgroup analysis {\mob}, do not produce cluster labels, we further use {\gmm} to cluster the estimated treatment effects to obtain cluster labels.
We see that our {\basiccs} performs the best on the simulated dataset in terms of the metrics. All other supervised learning methods tend to either underestimate or overestimate the treatment effects, resulting in larger errors. Some individuals are misclassified based on treatment effects only as shown in Figure \ref{fig:HTEclustXsim1200 res}. For subgroup analysis method {\mob}, it will learn more clusters than the ground truth due to the fashion of splitting one covariate at a time, also poor in capturing interactions between multiple covariates.
Figure~\ref{fig:HTEclustXsim1200 UMAP extra} presents additional UMAP projections of covariate space colored by cluster assignments from {\itt}, {\findit}, and {\owe}. {\itt} identifies only $3$ clusters, merging subgroups with distinct treatment effects. {\findit} assigns nearly all individuals to a single dominant cluster, consistent with its near-degenerate SATE range and high \textsc{PEHE}. {\owe} recovers $4$ clusters but exhibits substantial misclassification across covariate regions, reflecting its limited ability to capture the underlying heterogeneity.

%\subsection{Prior Sensitivity Test}

\begin{table}[!hbt]
    \centering
    \caption{Sensitivity Test of Prior SD of constant $\beta_k$ on {\simHTE}.}\label{tab:sensitivity sd}
    \begin{tabular}{cccc}\\\toprule
         $\beta_k$ SD &  \textsc{ARI} & Range of $\tau_k$ & \textsc{PEHE}\\\midrule
        0.5 & 0.750 & [-4.810 4.984] & 0.847\\
        1 & 0.733 & [-4.766, 5.020] & 0.943\\
        5 & 0.733 & [-4.750, 4.927] & 0.944\\ 
        10 & 0.732 & [-4.841, 5.059] & 0.983\\
        100 & 0.739 & [-4.812, 5.006] & 0.987\\
        \bottomrule
    \end{tabular}
\end{table}

\begin{table}[!hbt]
    \centering
    \caption{Sensitivity Test of Prior Mean of constant $\beta_k$ on {\simHTE}.}\label{tab:sensitivity mean}
    \begin{tabular}{cccc}\\\toprule
         $\beta_k$ Mean &  \textsc{ARI} & Range of $\tau_k$ & \textsc{PEHE}\\\midrule
        (0,0,0,0) & 0.733 & [-4.766, 5.020] & 0.943\\
        (0,0,-5,5) & 0.751 & [-4.790, 5.029] & 0.885\\
        \bottomrule
    \end{tabular}
\end{table}

\subsection{Homogeneous and Null Treatment Effects Test.}
We are clustering over covariates and treatment outcomes simultaneously. Thus, in the homogeneous and null treatment effects case, we can find clusters over covariates as a standard Bayesian mixture model will, but without differences in treatment effects, or we can find clusters with similar covariates but opposite treatment effects, as illustrated by the results in Table \ref{tab:homo null TE}. In the test set, however, these might lead to null treatment effects clusters with redundant or tiny clusters, showing some inconsistency between training and test results. Only when we can choose the correct number of clusters by validation metrics to be 1 can we avoid the false positive discovery.

\begin{table}[!hbt]
    \centering
    \caption{Test of Homogeneous and Null Treatment Effects on {\simNull}. We have only one cluster of individuals with zero treatment effects.}\label{tab:homo null TE}
    \begin{tabular}{ccccccc}\\\toprule
         K & Train RMSE  &  Test RMSE & Train \textsc{PEHE} & Test \textsc{PEHE} & Train Range & Test Range \\
         & of Control & of Control & & & of $\tau_k$ & of SATE\\\midrule
        1 & 0.217 & 0.460 & 0.048 & 0.048 & [0.048, 0.048] & [-0.057,-0.057]\\ % 0.211, 0.525 0.044 0.044
        2 & 0.210 & 0.460 & 0.386 & 0.382 & [-0.347, 0.429] & [-0.172, -0.025]\\ % 0.204, 0.525 0.395 0.39
        3 & 0.201 & 0.529 & 0.386 & 0.268 & [-0.550, 0.733] & [-6.29, -0.048]\\ % 0.187. 0.527 0.376 0.195 / -0.114 first
        4 & 0.203 & 0.524 & 0.368 & 0.223 & [-0.694, 1.232] & [-0.138,0.029]\\ % 0.224 0.629 0.391 0.257
        5 & 0.216 & 0.472 & 0.361 & 0.189 & [-0.748, 1.240] & [-0.0224,-0.0215]\\ % 0.231 0.545 0.396 0.309
        \bottomrule
    \end{tabular}
\end{table}

\subsection{Posterior Distribution of Parameters}\label{app:posterior}
%The treatment effect is assumed to be the same for all members of populations characterized by a vector of observable variables $\bfx_n$. As a result, the (conditional) distribution of individual treatment effects $\tau_n$ is degenerate \citep{Firpo2019TreatmentDist}. This means $\p(\tau_n=\E[\tau_n\g\bfx_n,z_n])=1$.
%However, since we do not observe treatment effects $\tau_n$, we need to estimate them through the observed outcomes by linking to potential outcomes, that is
%\begin{align*}
%\tau(\bfx_n;\bfbeta_k)=\E[\tau_n\g\bfx_n,z_n]&=\E[y^1_n-y^0_n\g\bfx_n,z_n]=\E[y^1_n\g\bfx_n,z_n]-\E[y^0_n\g\bfx_n,z_n]\\
%&= \E[y_n^\obs\g a_n=1,\bfx_n,z_n]-\E[y_n^\obs\g a_n=0,\bfx_n,z_n] \tag{Unconfoundedness}
%\end{align*}
%which are identifiable given the data $\calD$ and correspond to the super-population causal parameters \citep{Ding2018CIBayes}. We hence further model the marginal distributions of potential outcomes:
Denote the parameters for potential outcomes as $\bftheta^y=\{\sigma_0,\sigma_1,\{\mu^0_n\}_{n=1}^N,\bfphi=\{\alpha,\rho\}\}$ where $\alpha,\rho$ are only available for Gaussian kernel.
Then the posterior of parameters $\bfTheta=\{\{\bftheta_k,\bfgamma_k,\bfbeta_k\}_{k=1}^K,\bftheta^y,\bfpi\}$ is:
\begin{align*}
    p(\bfTheta\g\calD) &\propto p(\calD\g\bfTheta)p(\bfTheta)\\
    & \propto \prod_{n=1}^N\left\{p(a_n\g\bfx_n)\sum_{k=1}^K\p(z_n=k)p(\bfx_n\g z_n=k)p(y_n^\obs\g a_n,\bfx_n,z_n=k)\right\}p(\bfTheta)\\
    &\propto \prod_{n=1}^N\left\{\sum_{k=1}^K\p(z_n=k)p(\bfx_n\g z_n=k)p(y_n^\obs\g a_n,\bfx_n,z_n=k)\right\}p(\bfTheta) \tag{$p(a_n\g\bfx_n)$ is known under randomization}\\
    &\propto \prod_{n=1}^N\left\{\sum_{k=1}^K\pi_kp(\bfx_n\g\bftheta_k)p(y_n^\obs\g a_n,\{\mu_n^0\}_{n=1}^N,\sigma_0,\sigma_1,\bfbeta_k)\right\}p(\bfTheta)
\end{align*}

\section{Additional Real Application Results}\label{app:real res}

\begin{figure}[!ht]
    \centering
    \includegraphics[width=0.5\linewidth]{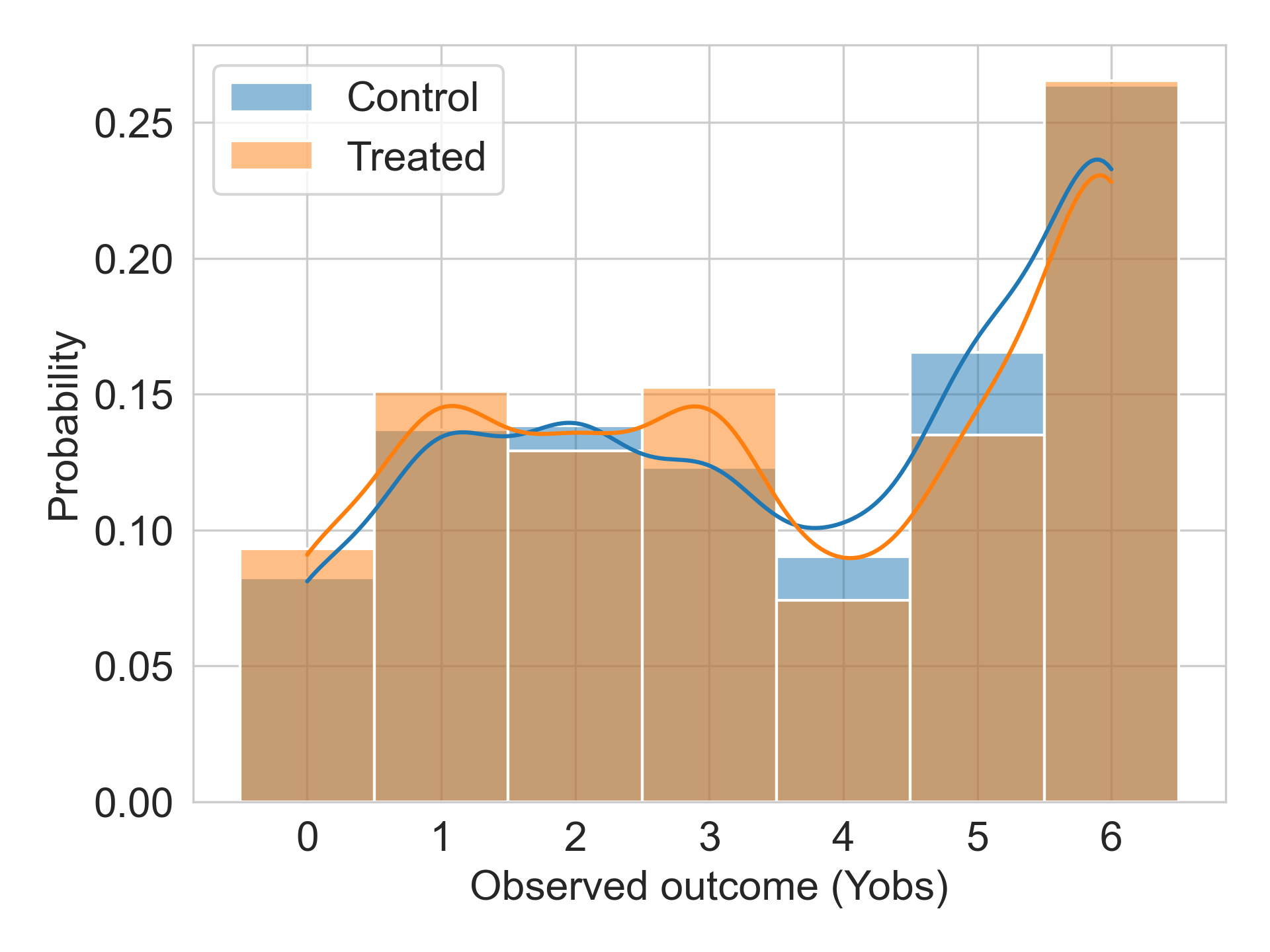}
    \caption{IST-3 Data: Oxford Handicap Score (OHS) at 6 months outcome distributions for control and treated groups.}
    \label{fig:IST3_norm-Y distribution}
\end{figure}

\begin{figure}[!ht]
    \centering
    \begin{subfigure}[c]{0.49\linewidth}
\includegraphics[width=\linewidth]{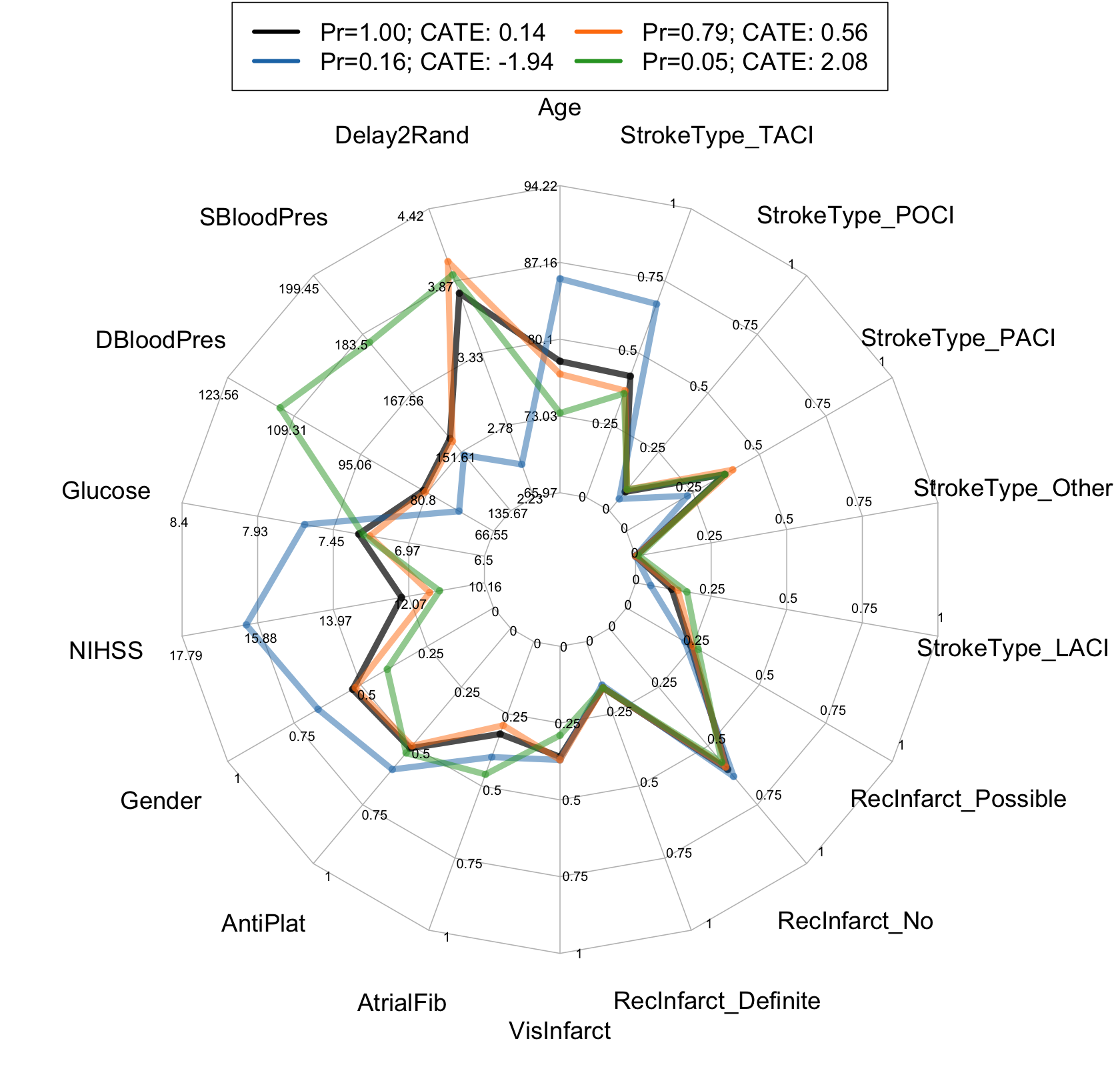}
  \caption{Estimated cluster means on train set}
    \end{subfigure}
    \hfill
    \begin{subfigure}[c]{0.49\linewidth}
        \includegraphics[width=\linewidth]{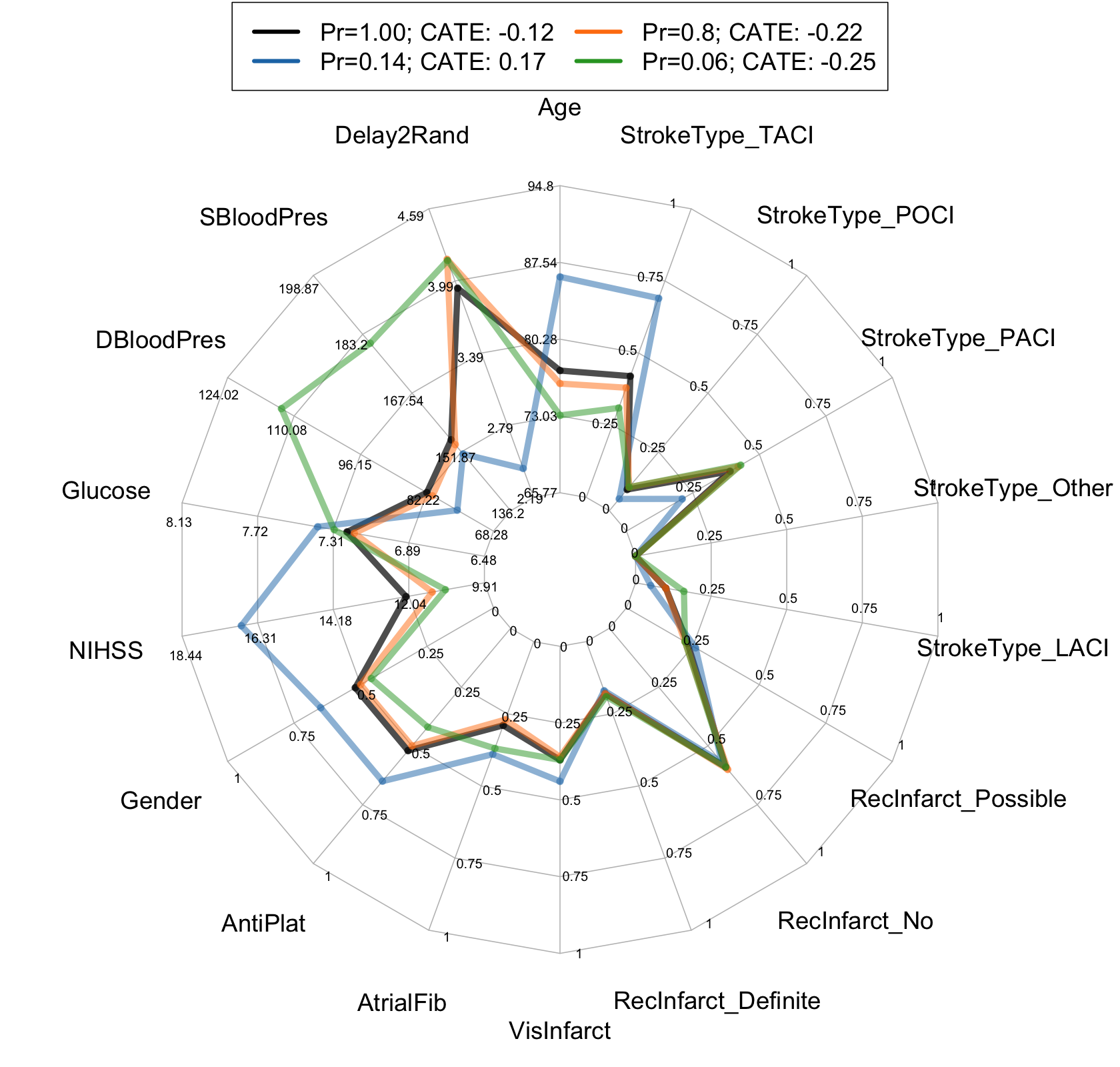}
         \caption{Empirical cluster means on test set}
    \end{subfigure}
    \\
    \begin{subfigure}[c]{0.49\linewidth}
        \includegraphics[width=\linewidth]{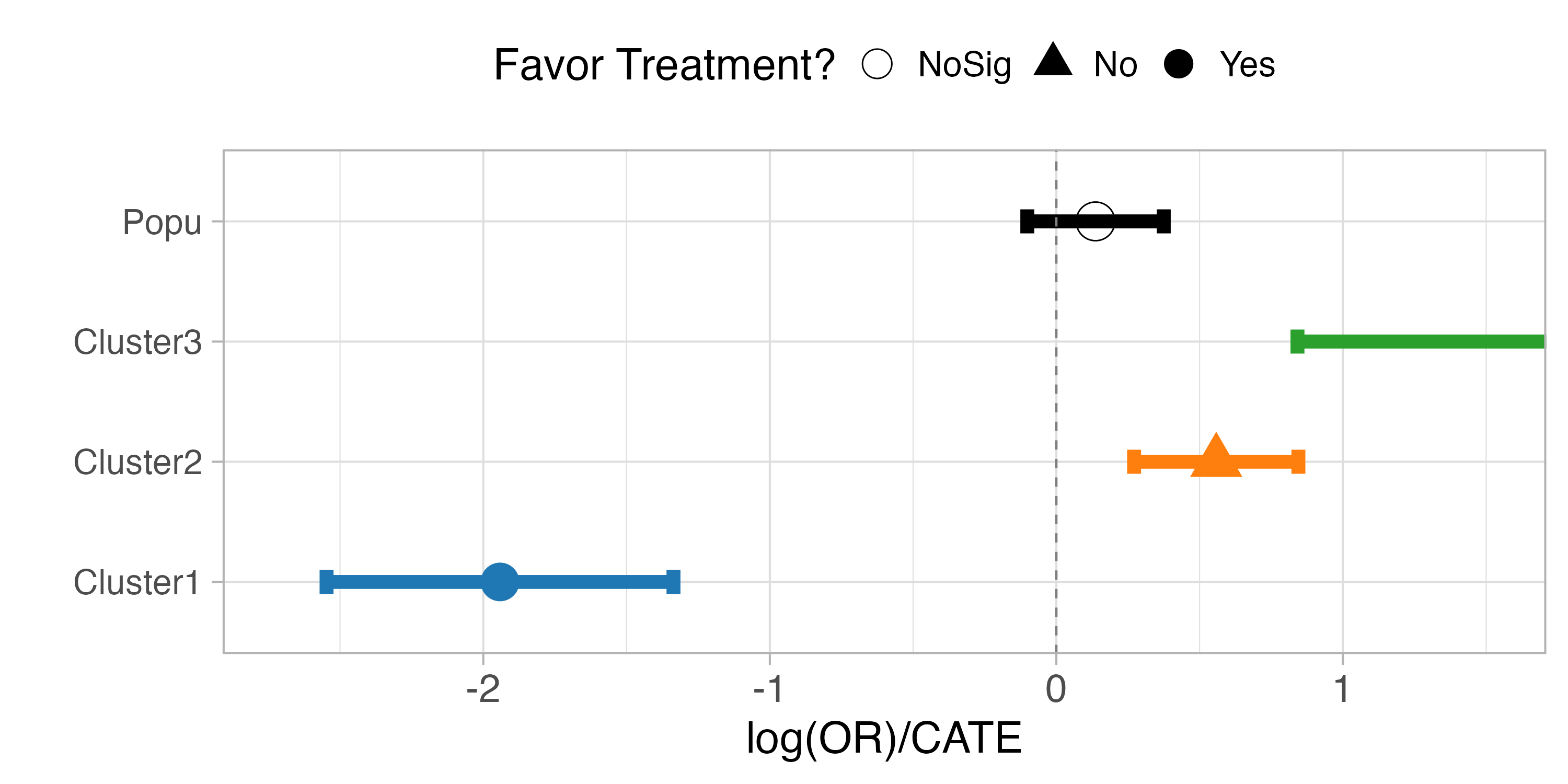}
             \caption{ORs with $95\%$ CIs for train set}
    \end{subfigure}
    \hfill
    \begin{subfigure}[c]{0.49\linewidth}
        \includegraphics[width=\linewidth]{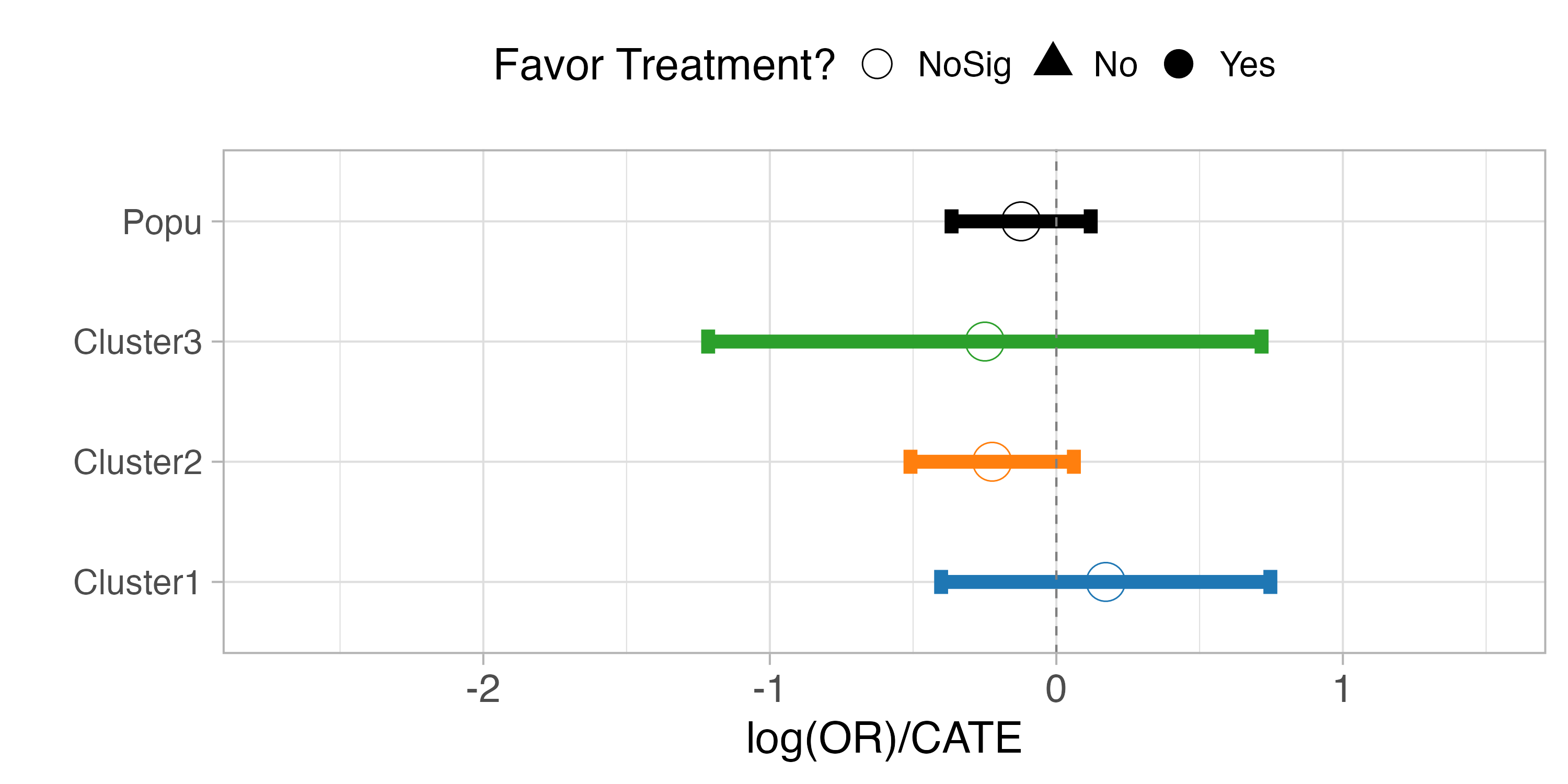}
        \caption{ORs with $95\%$ CIs for test set}
    \end{subfigure}
    \caption{{\cf} on IST-3 Dataset. }
    \label{fig:IST3 CF res}
\end{figure}

\begin{figure}[!ht]
    \centering
    \begin{subfigure}[c]{0.49\linewidth}
\includegraphics[width=\linewidth]{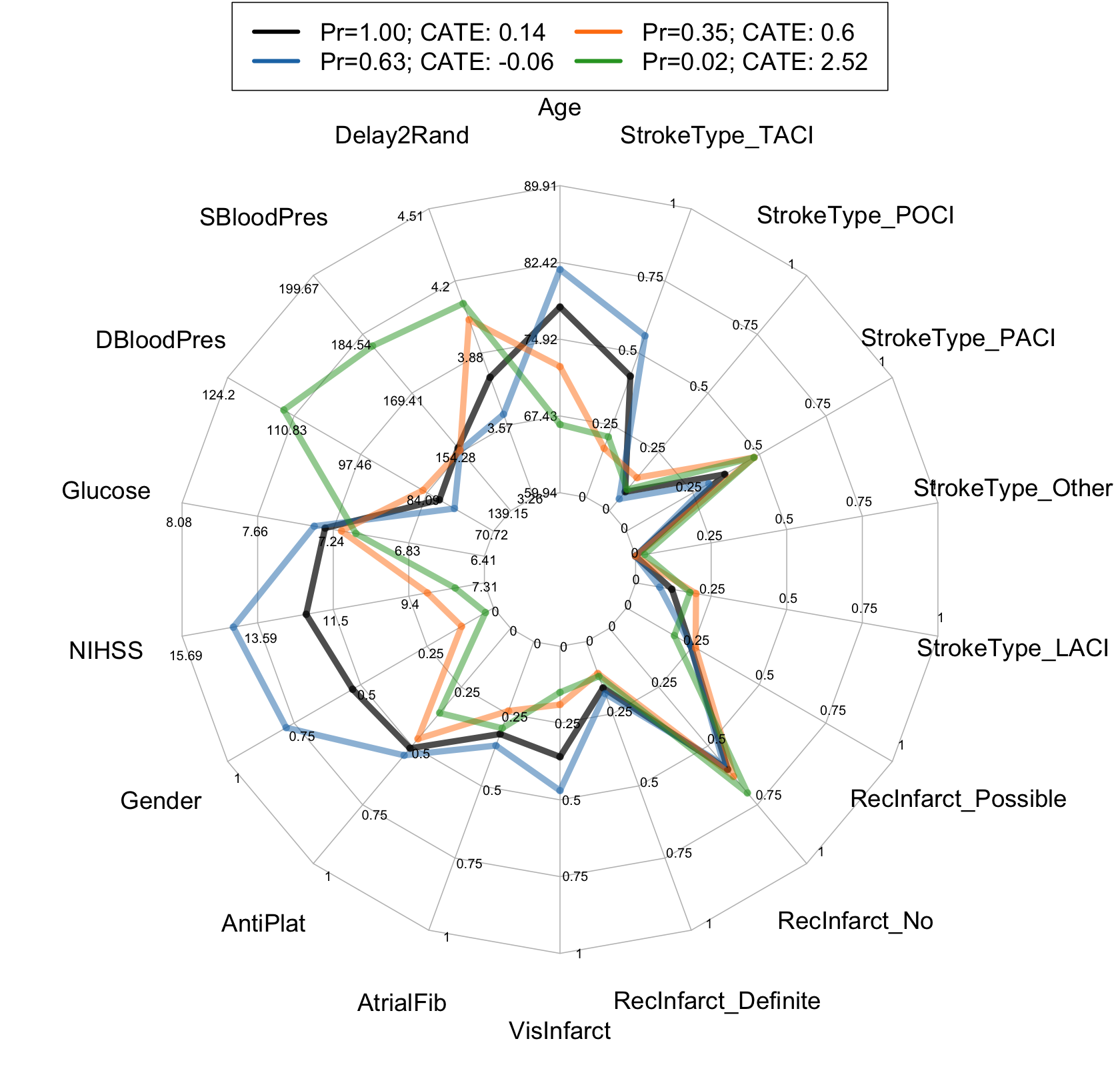}
  \caption{Estimated cluster means on train set}
    \end{subfigure}
    \hfill
    \begin{subfigure}[c]{0.49\linewidth}
        \includegraphics[width=\linewidth]{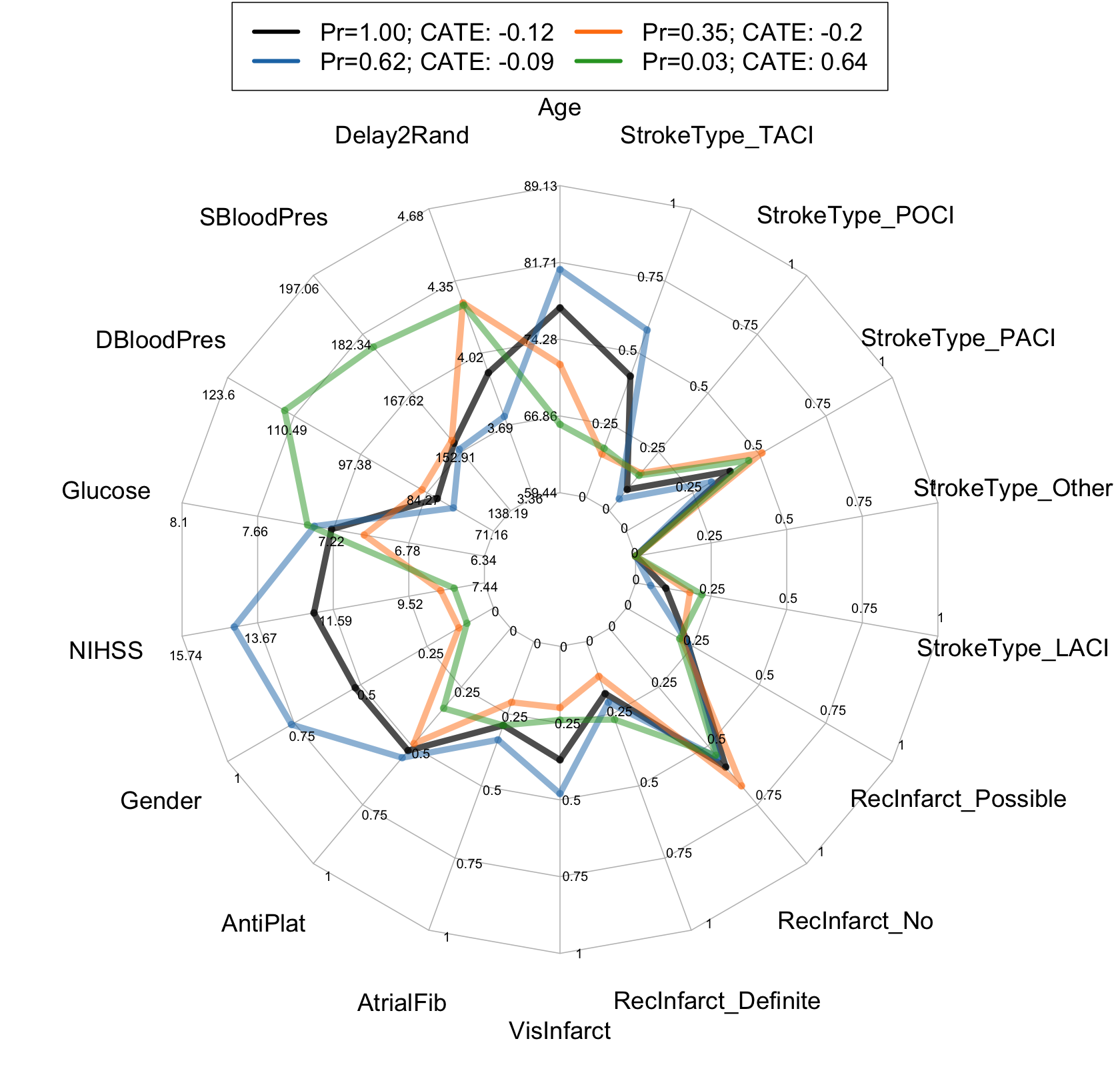}
         \caption{Empirical cluster means on test set}
    \end{subfigure}
    \\
    \begin{subfigure}[c]{0.49\linewidth}
        \includegraphics[width=\linewidth]{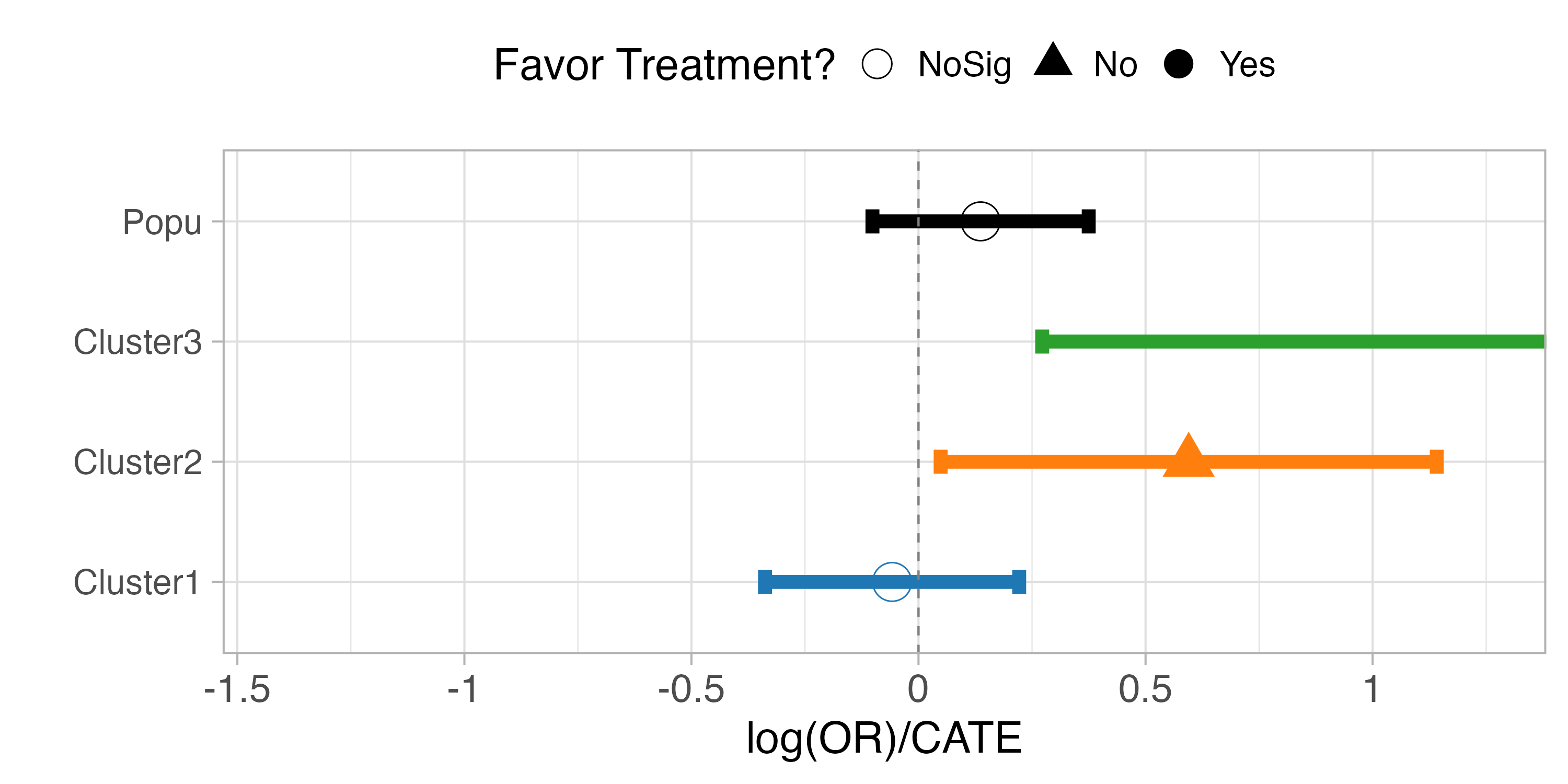}
             \caption{ORs with $95\%$ CIs for train set}
    \end{subfigure}
    \hfill
    \begin{subfigure}[c]{0.49\linewidth}
        \includegraphics[width=\linewidth]{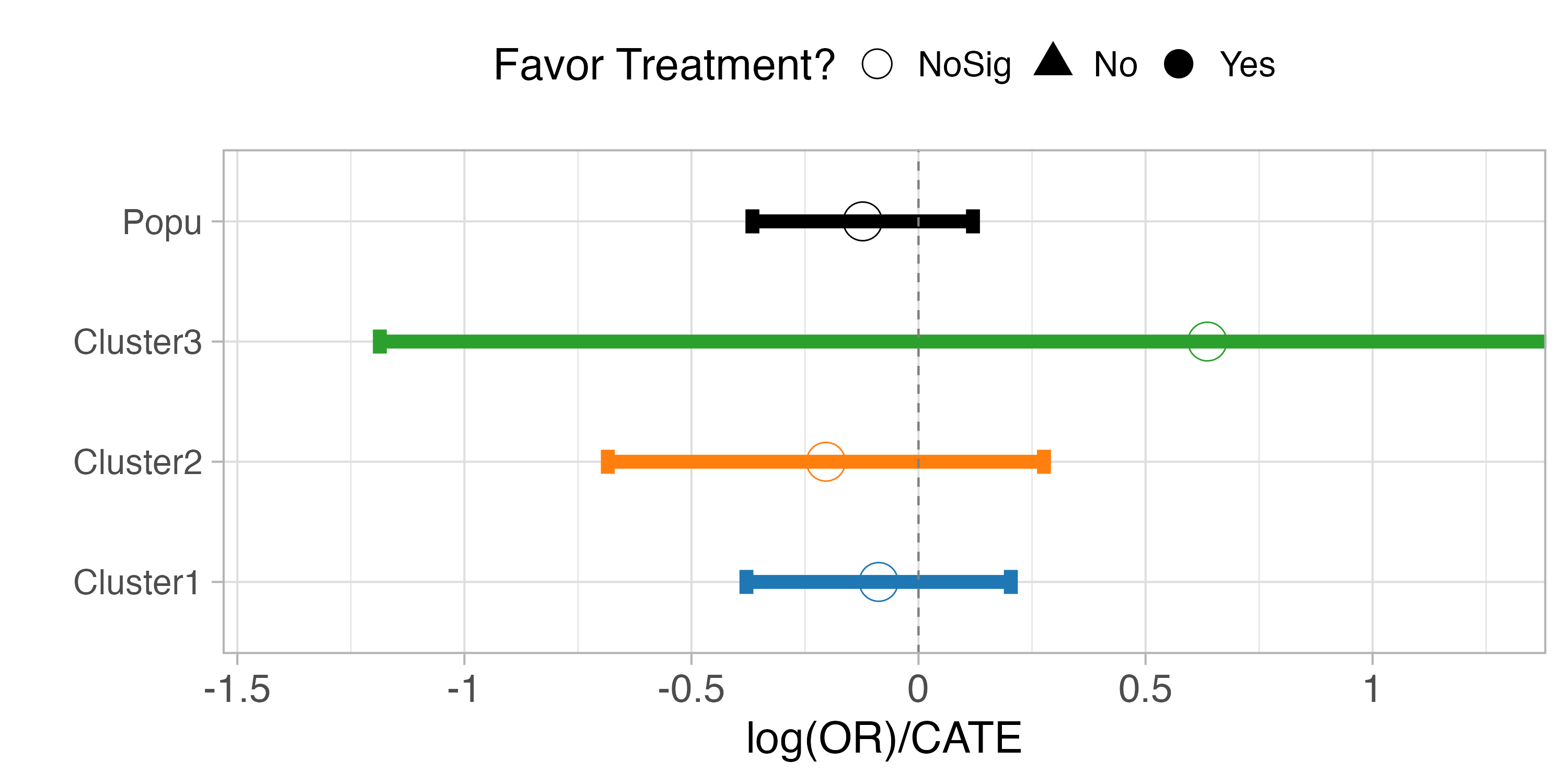}
        \caption{ORs with $95\%$ CIs for test set}
    \end{subfigure}
    \caption{{\bart} on IST-3 Dataset. }
    \label{fig:IST3 BART res}
\end{figure}

\begin{figure}[!ht]
    \centering
    \begin{subfigure}[c]{0.49\linewidth}
\includegraphics[width=\linewidth]{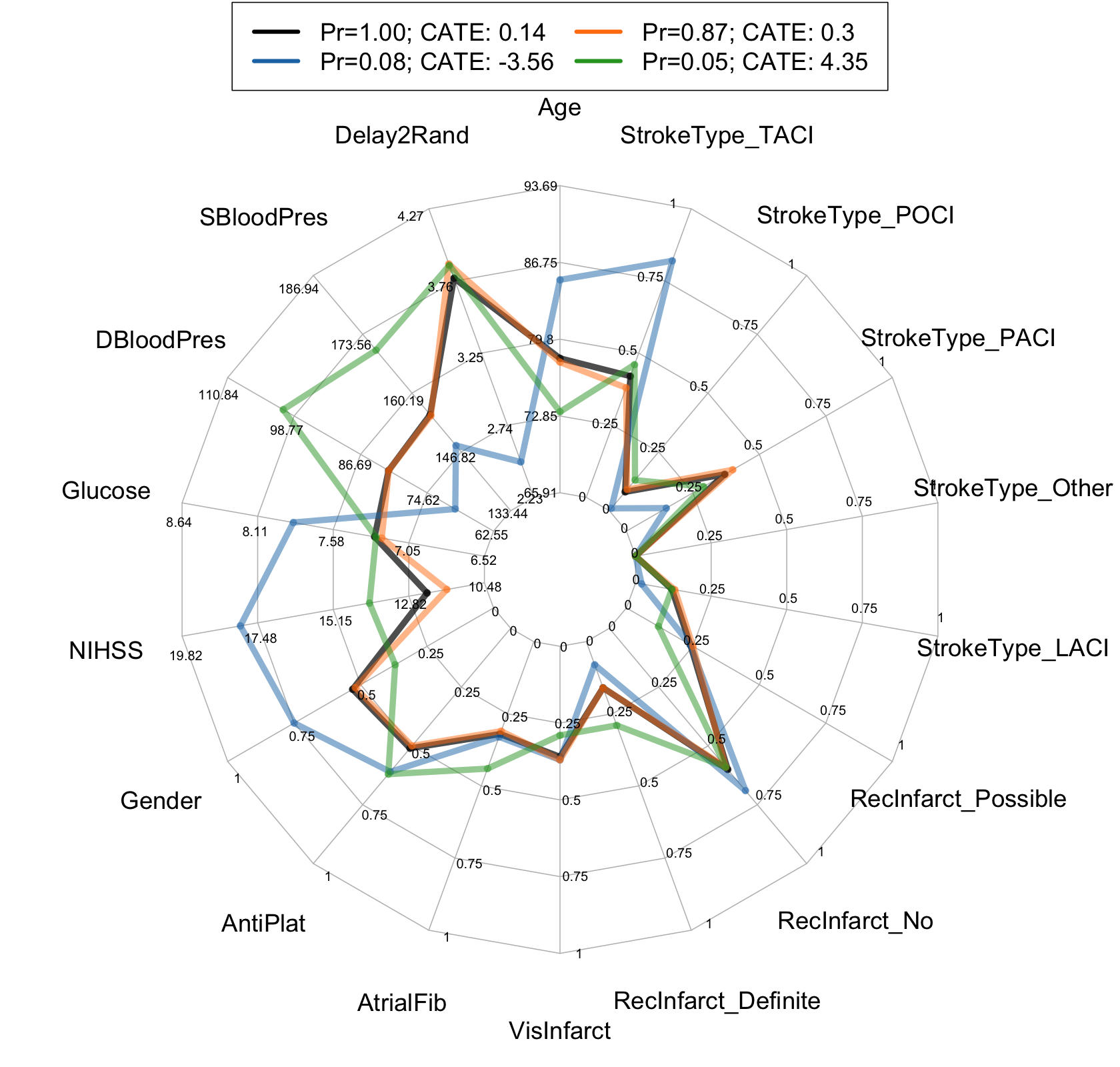}
  \caption{Estimated cluster means on train set}
    \end{subfigure}
    \hfill
    \begin{subfigure}[c]{0.49\linewidth}
        \includegraphics[width=\linewidth]{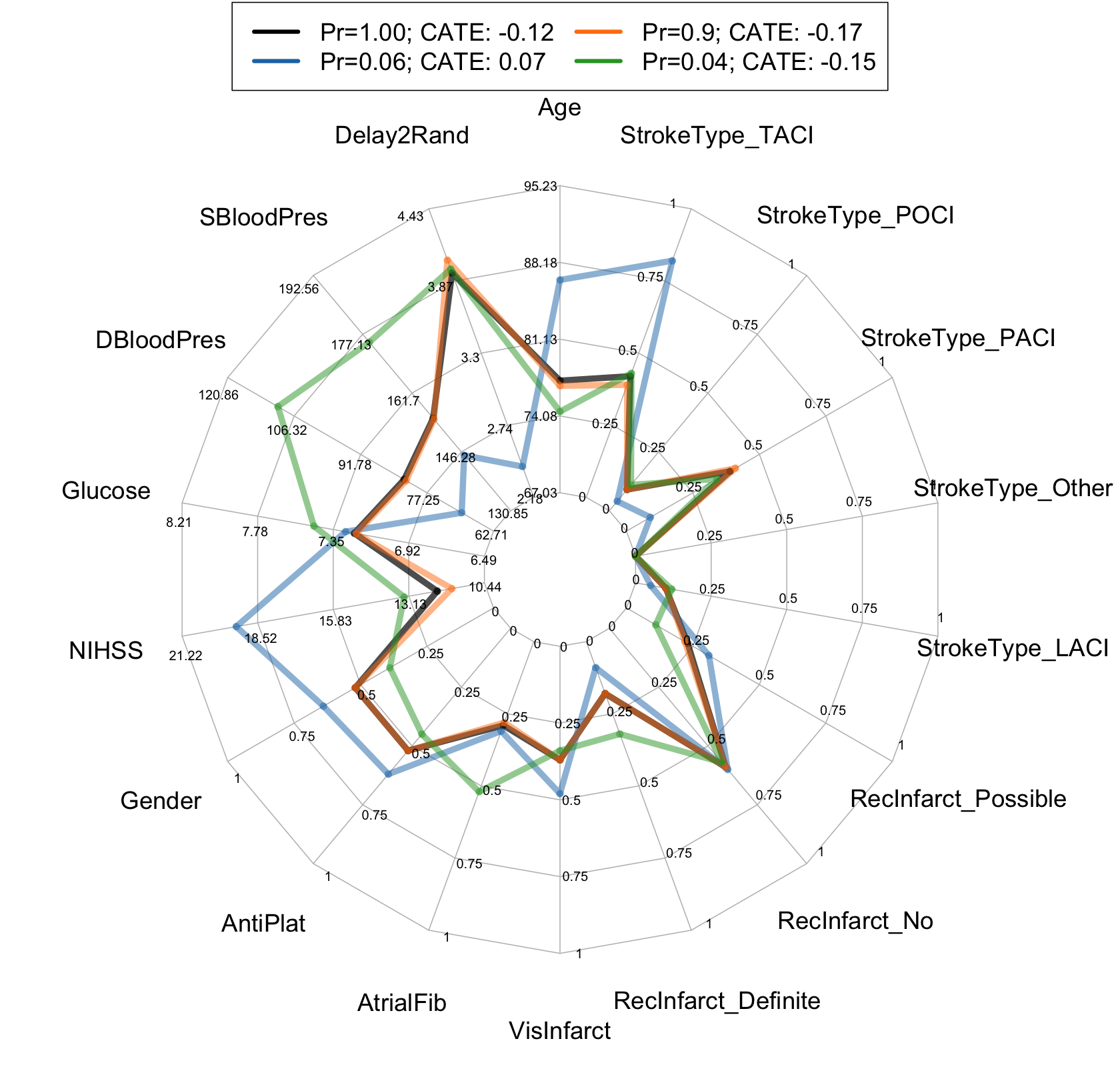}
         \caption{Empirical cluster means on test set}
    \end{subfigure}
    \\
    \begin{subfigure}[c]{0.49\linewidth}
        \includegraphics[width=\linewidth]{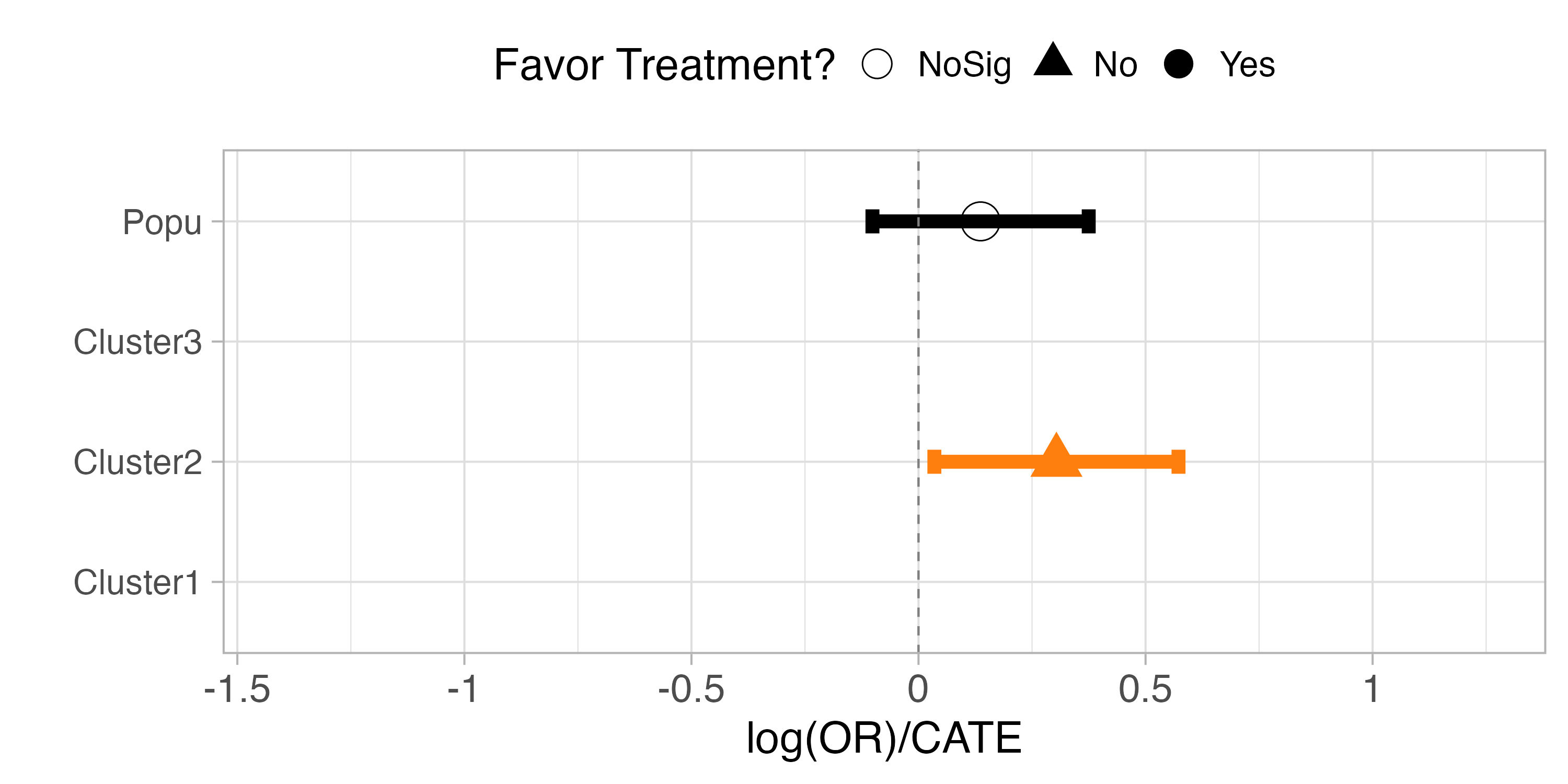}
             \caption{ORs with $95\%$ CIs for train set}
    \end{subfigure}
    \hfill
    \begin{subfigure}[c]{0.49\linewidth}
        \includegraphics[width=\linewidth]{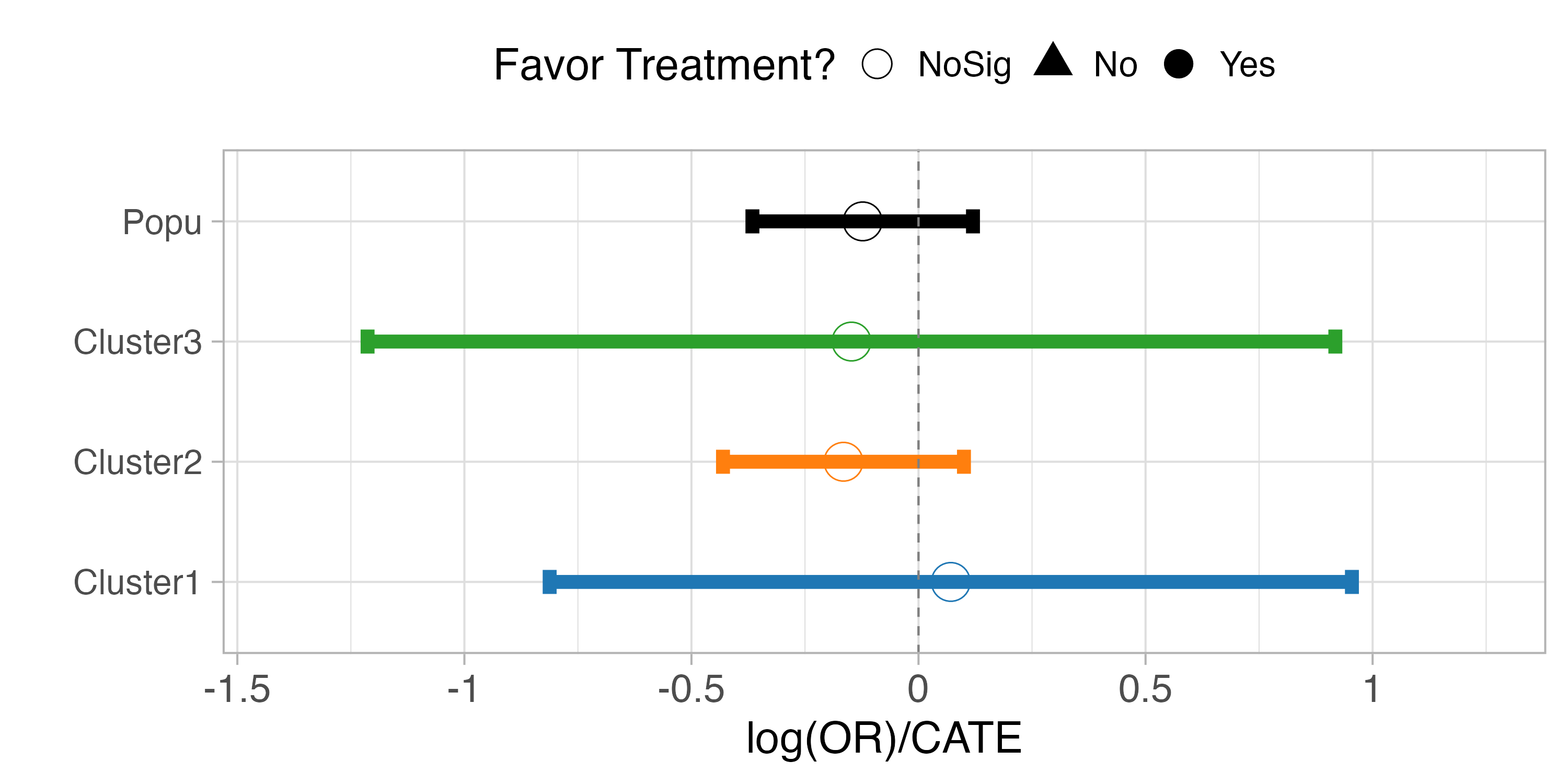}
        \caption{ORs with $95\%$ CIs for test set}
    \end{subfigure}
    \caption{\textsc{DR-learner} on IST-3 Dataset. }
    \label{fig:IST3 DRlearn res}
\end{figure}

\begin{figure}[!ht]
    \centering
    \begin{subfigure}[c]{0.49\linewidth}
\includegraphics[width=\linewidth]{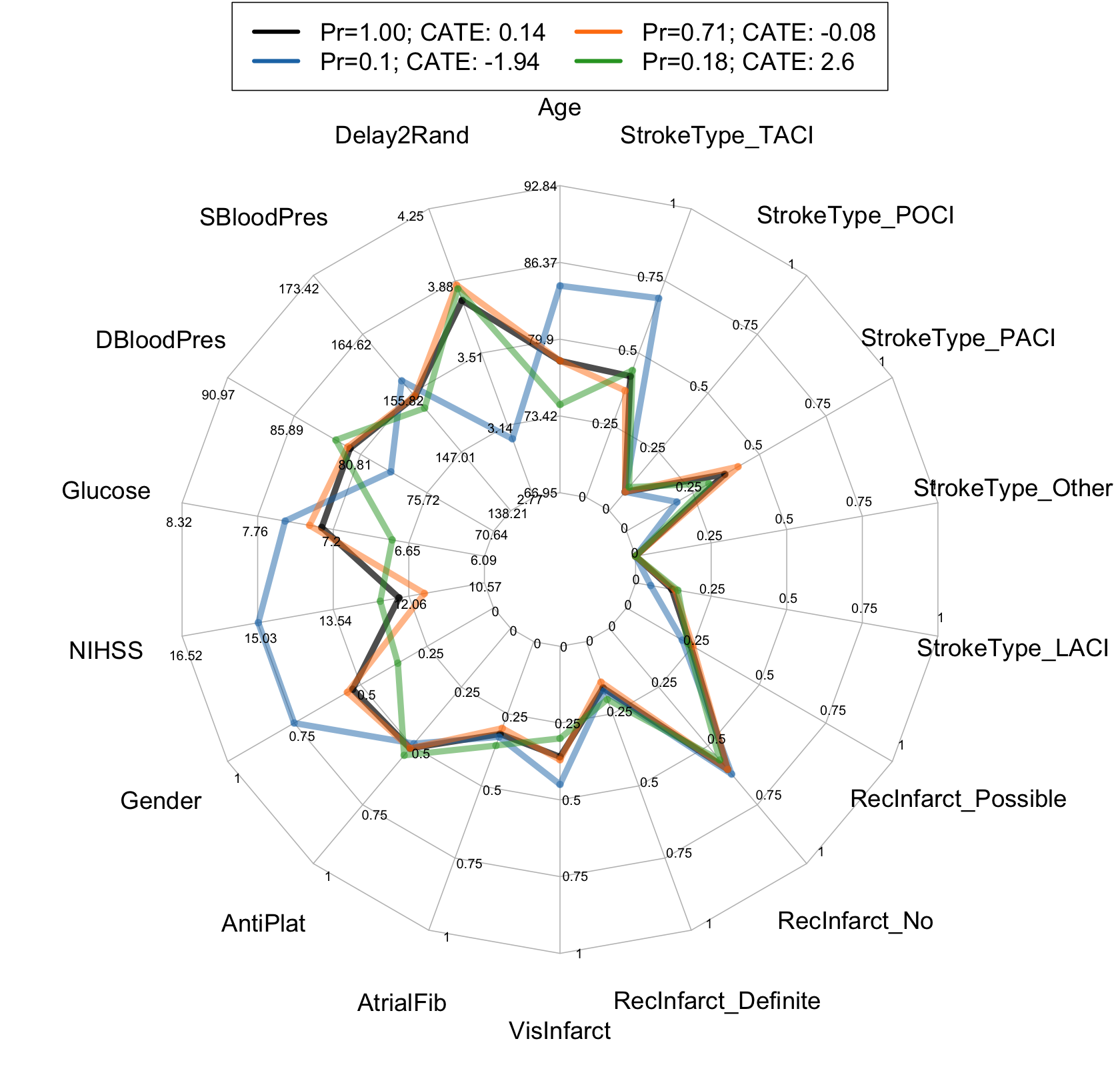}
  \caption{Estimated cluster means on train set}
    \end{subfigure}
    \hfill
    \begin{subfigure}[c]{0.49\linewidth}
        \includegraphics[width=\linewidth]{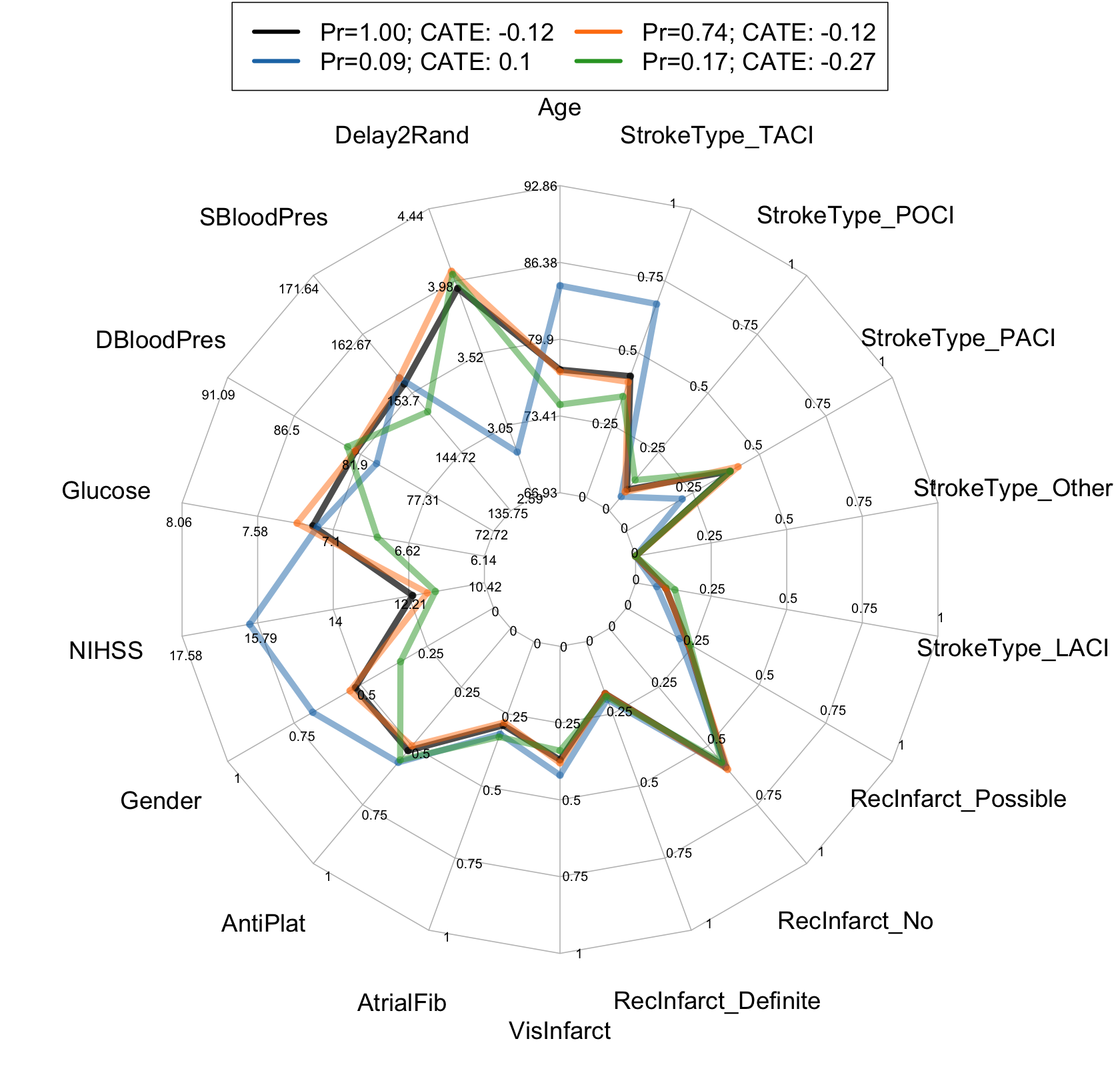}
         \caption{Empirical cluster means on test set}
    \end{subfigure}
    \\
    \begin{subfigure}[c]{0.49\linewidth}
        \includegraphics[width=\linewidth]{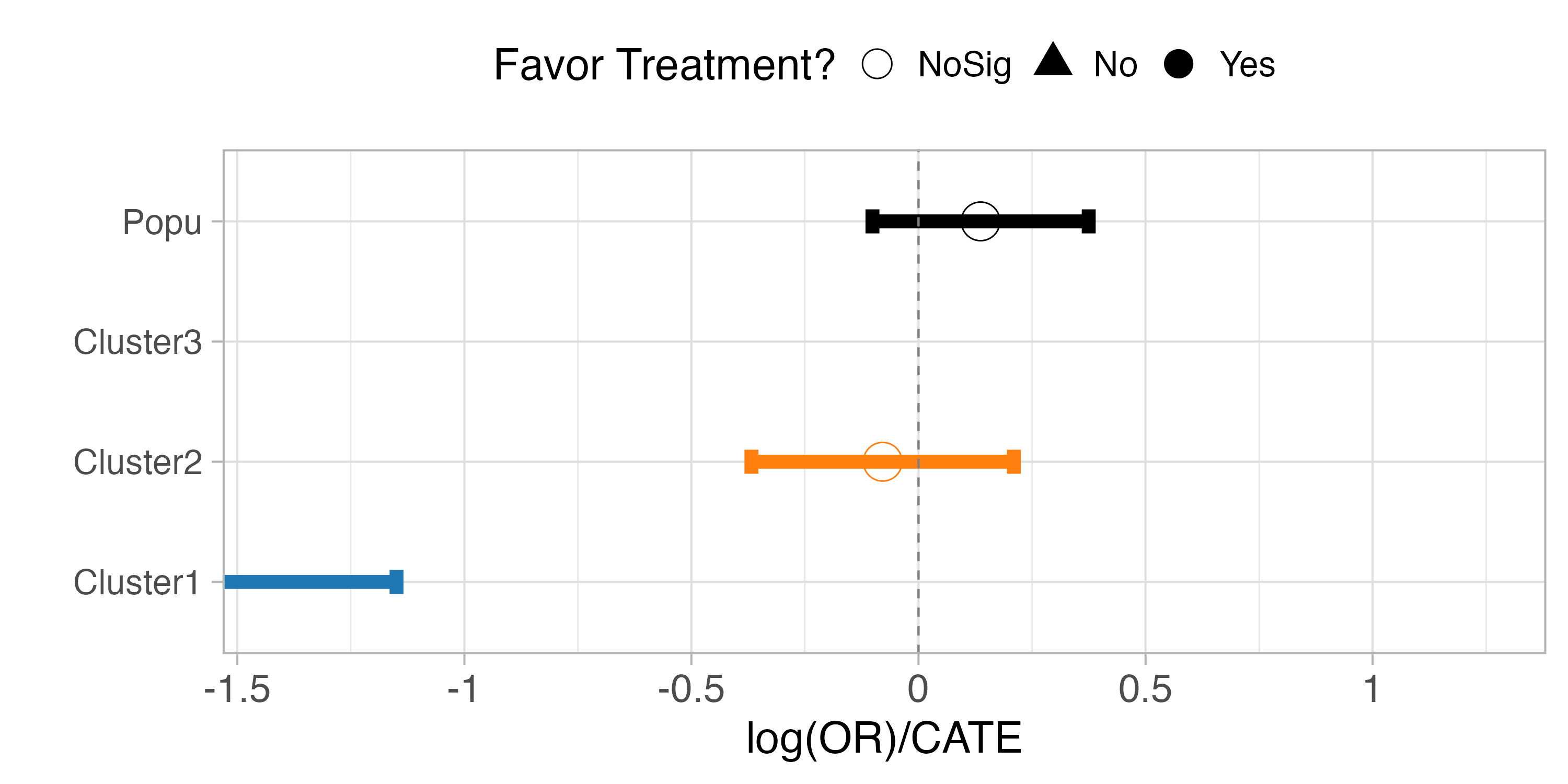}
             \caption{ORs with $95\%$ CIs for train set}
    \end{subfigure}
    \hfill
    \begin{subfigure}[c]{0.49\linewidth}
        \includegraphics[width=\linewidth]{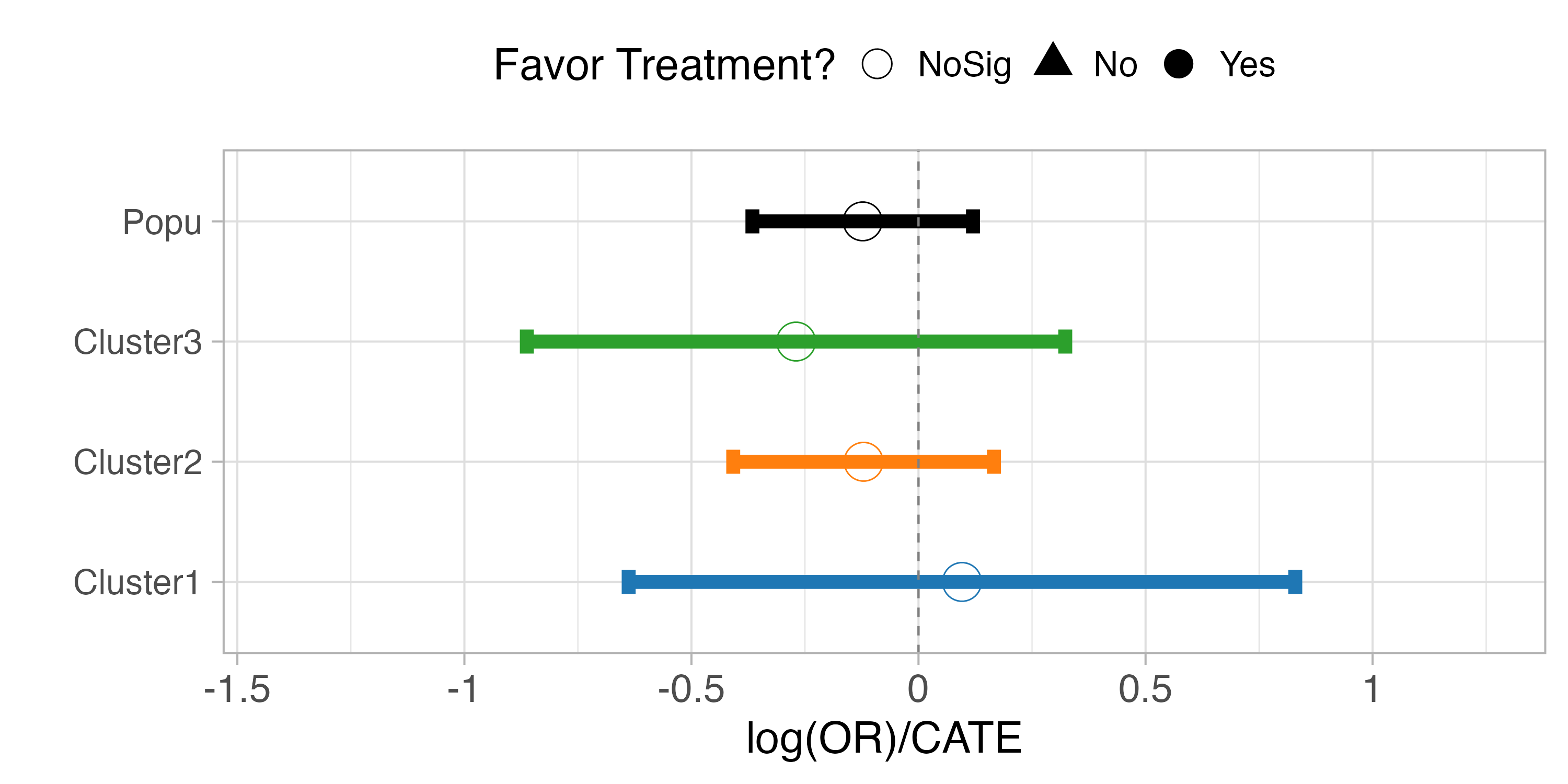}
        \caption{ORs with $95\%$ CIs for test set}
    \end{subfigure}
    \caption{\textsc{R-learner} on IST-3 Dataset. }
    \label{fig:IST3 Rlearn res}
\end{figure}

\begin{figure}[!ht]
    \centering
    \begin{subfigure}[c]{0.49\linewidth}
\includegraphics[width=\linewidth]{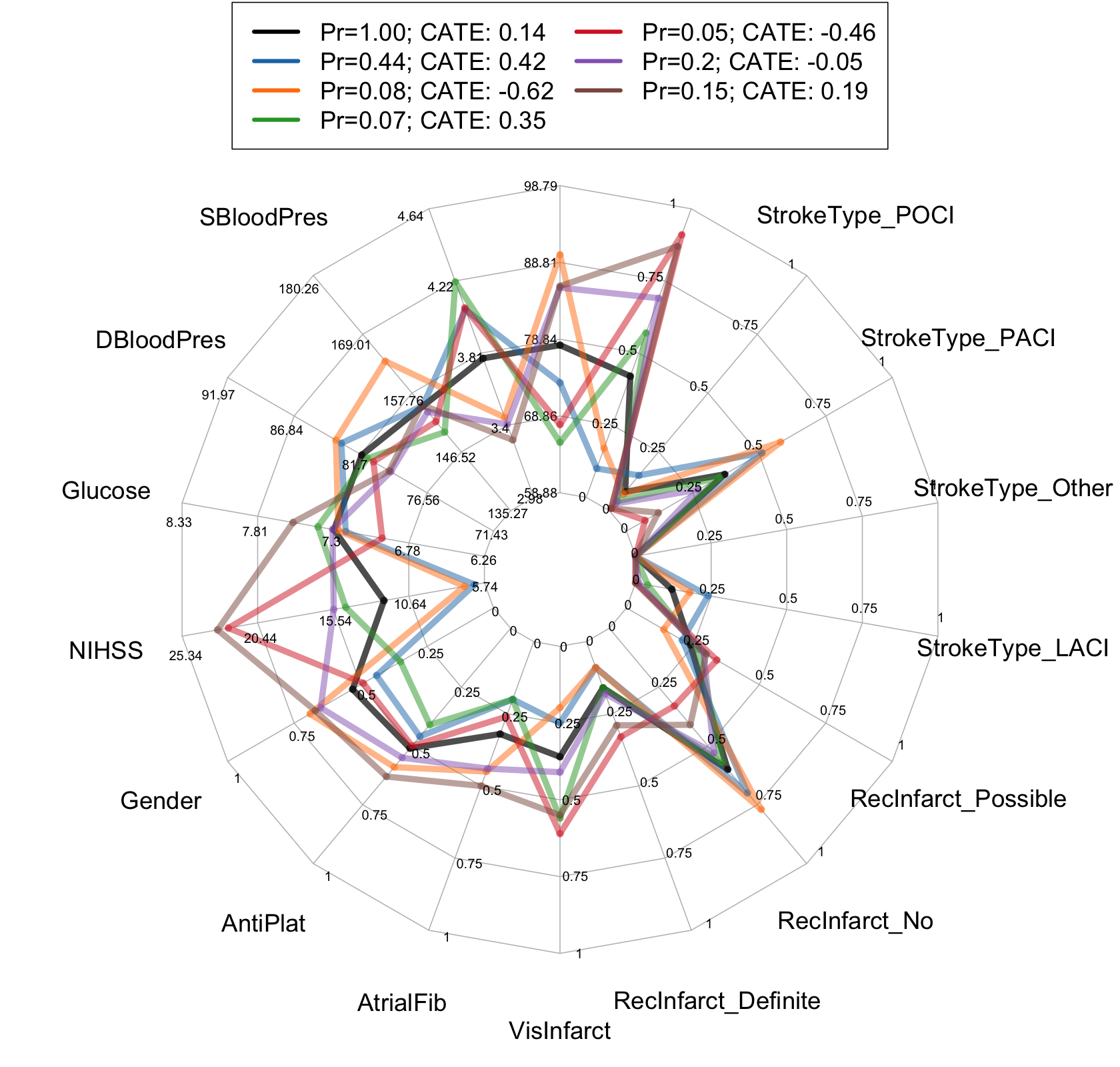}
  \caption{Estimated cluster means on train set}
    \end{subfigure}
    \hfill
    \begin{subfigure}[c]{0.49\linewidth}
        \includegraphics[width=\linewidth]{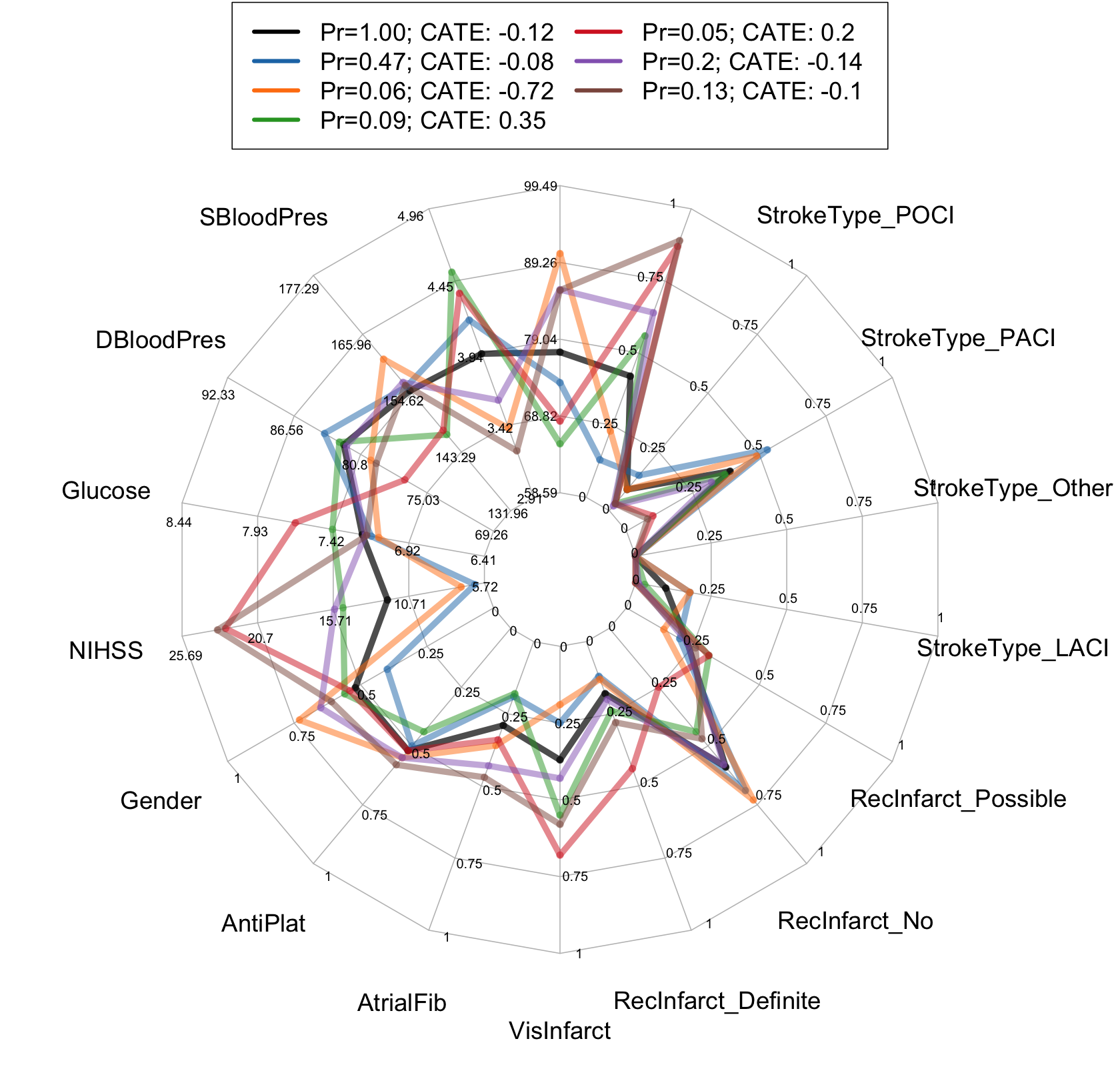}
         \caption{Empirical cluster means on test set}
    \end{subfigure}
    \\
    \begin{subfigure}[c]{0.49\linewidth}
        \includegraphics[width=\linewidth]{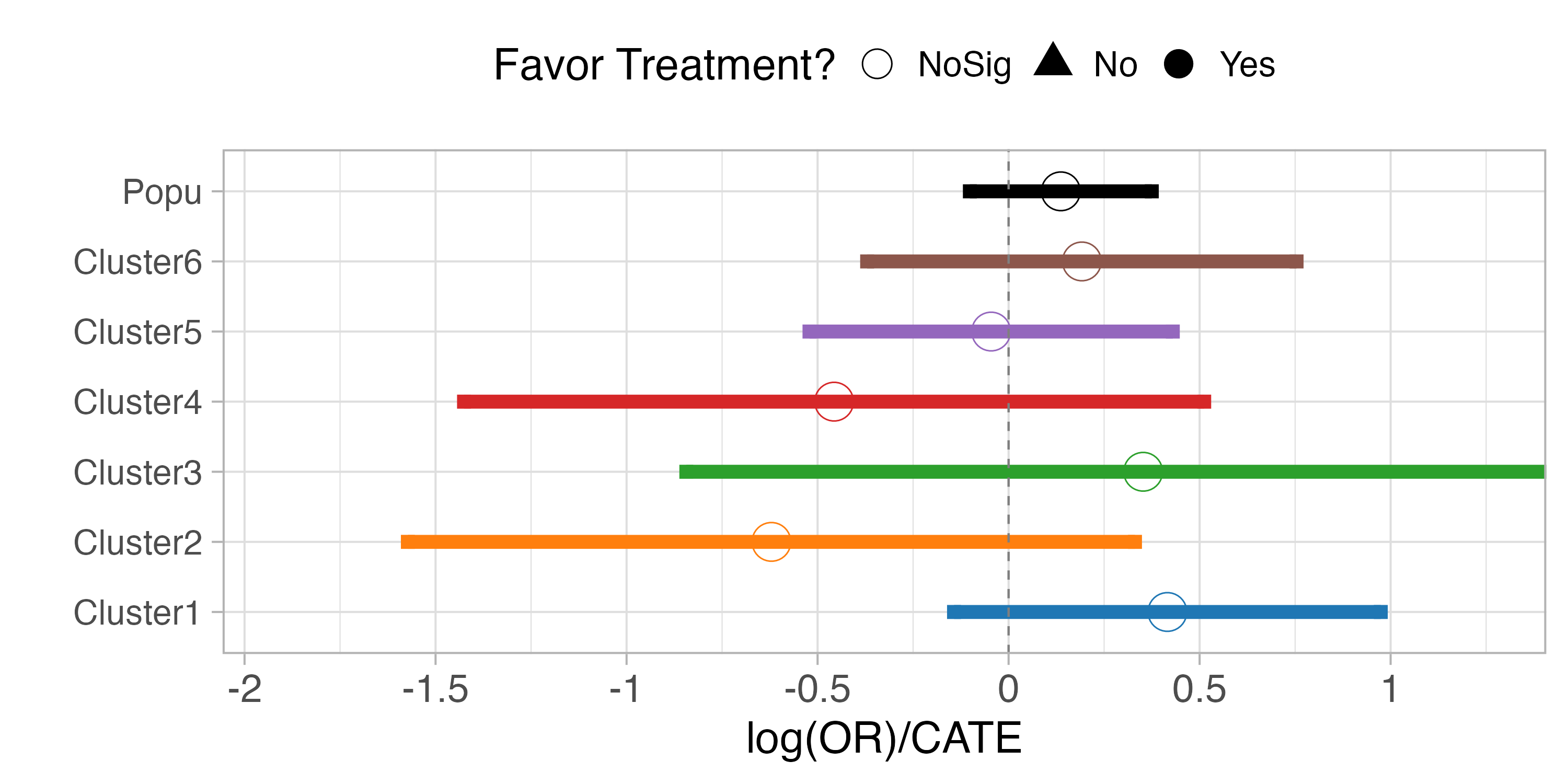}
             \caption{ORs with $95\%$ CIs for train set}
    \end{subfigure}
    \hfill
    \begin{subfigure}[c]{0.49\linewidth}
        \includegraphics[width=\linewidth]{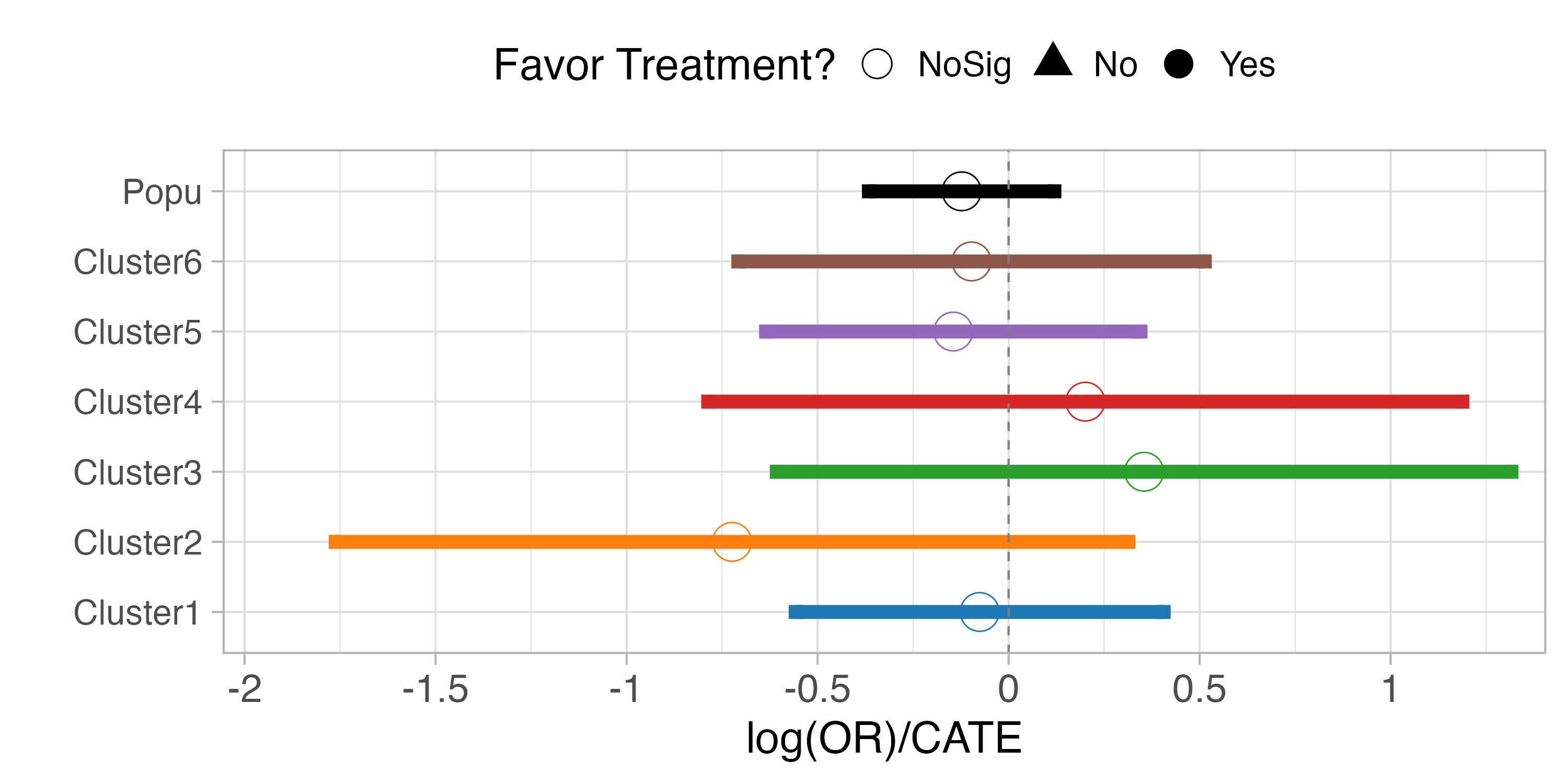}
        \caption{ORs with $95\%$ CIs for test set}
    \end{subfigure}
    \caption{{\mob} on IST-3 Dataset. }
    \label{fig:IST3 MOB res}
\end{figure}

Figure \ref{fig:IST3_norm-Y distribution} shows the distribution of OHS outcomes for the IST-3 dataset. Table \ref{tab:IST-3 list of covariates} describes the covariates selected for the IST-3 dataset. Table \ref{tab:cluster mean train} and Table \ref{tab:contingency table train} summarise the detailed statistics and resulting values for the training set corresponding to the cluster means radar chart and OR plots in the main content, while Table \ref{tab:cluster mean test} and Table \ref{tab:contingency table test} summarise those for the test set. Table \ref{tab:contingency table GMM} summarises the contingency table of clustering given by {\gmm}. Table \ref{tab:real dataset & sup res} presents the performance of {\basiccs} against other supervised learning models on the IST-3 dataset. We see that despite slightly smaller test policy risk (max difference around $0.03$, which is about $40$ individuals' decisions due to uncertainty) of supervised learning methods, all the supervised learning models yield unstable subgroups in terms of log OR (SATE) based solely on treatment effects, as shown by the train and test range of SATE. The corresponding radar charts and OR plots are shown in Figure \ref{fig:IST3 CF res}, Figure \ref{fig:IST3 BART res}, Figure \ref{fig:IST3 DRlearn res} and Figure \ref{fig:IST3 Rlearn res}. For the subgroup analysis method {\mob} in Figure \ref{fig:IST3 MOB res}, we see that it is relatively consistent in training and test results, although no significant clusters are found. However, it obtained four clusters based on the splitting covariates ``NIHSS" and ``age" only, which ignored important covariates such as stroke subtypes found by {\basiccs}.

\begin{table*}[!ht]
    \centering
    \small
     \caption{List of Covariates Used for the IST-3 Dataset.}
    \label{tab:IST-3 list of covariates}
    \begin{tabular}{ll}
    \toprule
         Description of Covariate& Feature Type\\\midrule
         Age& Continuous
\\
         Delay Time from onset of stroke symptoms to randomisation ($\si{\hour}$)& Continuous
\\
         Systolic blood pressure ($\si{\mm H\g}$)& Continuous
\\
         Diastolic blood pressure ($\si{\mm H\g}$)& Continuous
\\
         Glucose ($\si{\mmol/\L}$)& Continuous
\\
         NIHSS (Stroke Score at randomisation)& Continuous
\\
         Sex& Binary\\
         Treatment with antiplatelet drugs in previous $\qty{48}{\hour}$?& Binary\\
         Atrial fibrillation?& Binary\\
 Acute ischaemic change on randomisation scan according to  &Binary\\
 expert panel? & \\
 Clinician’s assessment of recent ischaemic change at randomisation &Categorical (No, Possible, Definite)\\
 Stroke syndrome (type)&Categorical \\
 & (TACI, PACI, LACI, POCI, Other)\\\bottomrule
    \end{tabular}
\end{table*}

\begin{table}[!ht]
    \centering
    \caption{The Evaluation Metrics for Selecting the Number of Clusters of the IST-3 Dataset.}
    \label{tab:IST-3 select the number of clusters}
    \begin{tabular}{ccc} \toprule
         No. Clusters&  Test Acc. of & Test Policy Risk\\
         & Control Outcome &\\\midrule
         $K=2$& 0.754 & 0.648\\
         $K=3$& \textbf{0.770} & \textbf{0.644}\\
         $K=4$& 0.762 & 0.644 \\\bottomrule
    \end{tabular}
\end{table}

\begin{table}[!hbt]
    \centering
    \caption{A Summary of Model Performance for Different Supervised Learning Models on IST-3 Data.}\label{tab:real dataset & sup res}
    \begin{tabular}{cccc}\\\toprule
         Model Case & Train Range of SATE & Test Range of SATE & Test Policy Risk\\\midrule
        {\basiccs} & [-0.273, 0.658] & [-0.243, 0.501] & 0.644\\
        {\cf} & [-1.942, 2.079] & [-0.249, 0.172] & 0.616 \\
        {\bart} & [-0.058, 2.521] & [-0.204, 0.636] & 0.632 \\
        \textsc{DR-learner} & [-3.563, 4.348] & [-0.165, 0.071] & 0.621 \\
        \textsc{R-learner} & [-1.941, 2.603] & [-0.270, 0.096] & 0.618 \\
        {\mob} & [-0.621, 0.416] & [-0.724, 0.355] & 0.638 \\
        \bottomrule
    \end{tabular}
\end{table}

\begin{table*}[!htb]
\centering
\caption{IST-3: Summary Statistics (Min/Max/Mean) of Covariates by Clusters for Training Set ({\basiccs}).}
\label{tab:cluster mean train}
\begin{tabular}{lcccccc}
\toprule
  & Max & Min & Popu & Cluster1 & Cluster2 & Cluster3\\
\midrule
Age & 89.84 & 66.08 & 78.07 & 73.42 & 81.68 & 80.54\\
Delay2Rand & 4.37 & 3.14 & 3.78 & 3.98 & 3.49 & 3.68\\
SBloodPres & 171.79 & 139.70 & 155.27 & 155.22 & 155.38 & 156.17\\
DBloodPres & 92.70 & 71.42 & 81.61 & 84.28 & 79.36 & 82.03\\
Glucose & 8.06 & 6.30 & 7.29 & 7.32 & 7.32 & 7.00\\
NIHSS & 19.50 & 7.31 & 12.25 & 8.12 & 17.73 & 12.22\\
\addlinespace
Gender & 1.00 & 0.00 & 0.53 & 0.42 & 0.54 & 0.61\\
AntiPlat & 1.00 & 0.00 & 0.51 & 0.51 & 0.52 & 0.51\\
AtrialFib & 1.00 & 0.00 & 0.32 & 0.25 & 0.38 & 0.32\\
VisInfarct & 1.00 & 0.00 & 0.36 & 0.28 & 0.49 & 0.36\\
\addlinespace
RecInfarct\_Definite & 1.00 & 0.00 & 0.16 & 0.09 & 0.24 & 0.21\\
RecInfarct\_No & 1.00 & 0.00 & 0.60 & 0.67 & 0.50 & 0.60\\
RecInfarct\_Possible & 1.00 & 0.00 & 0.24 & 0.23 & 0.31 & 0.24\\
\addlinespace
StrokeType\_LACI & 1.00 & 0.00 & 0.12 & 0.21 & 0.01 & 0.08\\
StrokeType\_Other & 1.00 & 0.00 & 0.00 & 0.00 & 0.00 & 0.01\\
StrokeType\_PACI & 1.00 & 0.00 & 0.37 & 0.62 & 0.00 & 0.36\\
StrokeType\_POCI & 1.00 & 0.00 & 0.08 & 0.12 & 0.00 & 0.07\\
StrokeType\_TACI & 1.00 & 0.00 & 0.42 & 0.01 & 1.00 & 0.43\\
\bottomrule
\end{tabular}
\end{table*}

\begin{table*}[!ht]
\centering
\caption{IST-3: Summary Statistics (Min/Max/Mean) of Covariates by Clusters for Test Set ({\basiccs}).}\label{tab:cluster mean test}
\begin{tabular}{lcccccc}
\toprule
  & Max & Min & Popu & Cluster1 & Cluster2 & Cluster3\\
\midrule
Age & 88.48 & 67.16 & 77.32 & 74.62 & 80.44 & 78.47\\
Delay2Rand & 4.52 & 3.34 & 3.93 & 4.10 & 3.71 & 3.88\\
SBloodPres & 171.26 & 138.54 & 155.09 & 155.69 & 153.93 & 155.61\\
DBloodPres & 91.99 & 72.21 & 82.26 & 83.63 & 80.23 & 82.42\\
Glucose & 8.07 & 6.30 & 7.23 & 7.25 & 7.34 & 7.00\\
NIHSS & 18.99 & 7.26 & 12.12 & 8.06 & 17.26 & 13.08\\
\addlinespace
Gender & 1.00 & 0.00 & 0.52 & 0.47 & 0.59 & 0.52\\
AntiPlat & 1.00 & 0.00 & 0.52 & 0.52 & 0.53 & 0.49\\
AtrialFib & 1.00 & 0.00 & 0.29 & 0.20 & 0.39 & 0.34\\
VisInfarct & 1.00 & 0.00 & 0.37 & 0.26 & 0.52 & 0.40\\
\addlinespace
RecInfarct\_Definite & 1.00 & 0.00 & 0.18 & 0.12 & 0.26 & 0.20\\
RecInfarct\_No & 1.00 & 0.00 & 0.59 & 0.69 & 0.48 & 0.56\\
RecInfarct\_Possible & 1.00 & 0.00 & 0.23 & 0.19 & 0.27 & 0.23\\
\addlinespace
StrokeType\_LACI & 1.00 & 0.00 & 0.10 & 0.18 & 0.00 & 0.09\\
StrokeType\_Other & 1.00 & 0.00 & 0.00 & 0.00 & 0.00 & 0.00\\
StrokeType\_PACI & 1.00 & 0.00 & 0.39 & 0.66 & 0.00 & 0.40\\
StrokeType\_POCI & 1.00 & 0.00 & 0.09 & 0.15 & 0.00 & 0.08\\
StrokeType\_TACI & 1.00 & 0.00 & 0.42 & 0.00 & 1.00 & 0.43\\
\bottomrule
\end{tabular}
\end{table*}

\begin{table*}[!ht]
\centering
\caption{IST-3: Contingency Table of Clusters for Training Set ({\basiccs}).}
\label{tab:contingency table train}
\begin{tabular}{ccccc}
\toprule
Cluster & Group & Mortality & Alive & Mortality Prop.\\
\midrule
Cluster1 & Treated & 53 & 265 & 16.7\%\\
Cluster1 & Control & 43 & 280 & 13.3\%\\
\addlinespace
Cluster2 & Treated & 92 & 133 & 40.9\%\\
Cluster2 & Control & 100 & 110 & 47.6\%\\
\addlinespace
Cluster3 & Treated & 49 & 86 & 36.3\%\\
Cluster3 & Control & 36 & 122 & 22.8\%\\
\bottomrule
\end{tabular}
\end{table*}

\begin{table*}[!ht]
\centering
\caption{IST-3: Contingency Table of Clusters for Test Set ({\basiccs}).}\label{tab:contingency table test}
\begin{tabular}{ccccc}
\toprule
Cluster & Group & Mortality & Alive & Mortality Prop.\\
\midrule
Cluster1 & Treated & 45 & 287 & 13.6\%\\
Cluster1 & Control & 51 & 255 & 16.7\%\\
\addlinespace
Cluster2 & Treated & 74 & 131 & 36.1\%\\
Cluster2 & Control & 102 & 143 & 41.6\%\\
\addlinespace
Cluster3 & Treated & 49 & 100 & 32.9\%\\
Cluster3 & Control & 30 & 101 & 22.9\%\\
\bottomrule
\end{tabular}
\end{table*}

\begin{table*}[!ht]
\centering
\caption{IST-3: Contingency Table of Clusters for {\gmm}.}\label{tab:contingency table GMM}
\begin{tabular}{ccccc}
\toprule
Cluster & Group & Mortality & Alive & Mortality Prop.\\
\midrule
Cluster1 & Treated & 244 & 276 & 46.9\%\\
Cluster1 & Control & 248 & 277 & 47.2\%\\
\addlinespace
Cluster2 & Treated & 81 & 488 & 14.2\%\\
Cluster2 & Control & 81 & 489 & 14.2\%\\
\addlinespace
Cluster3 & Treated & 37 & 238 & 13.5\%\\
Cluster3 & Control & 33 & 245 & 11.9\%\\
\bottomrule
\end{tabular}
\end{table*}

\begin{figure}[!ht]
  \centering
    \includegraphics[width=0.7\linewidth]{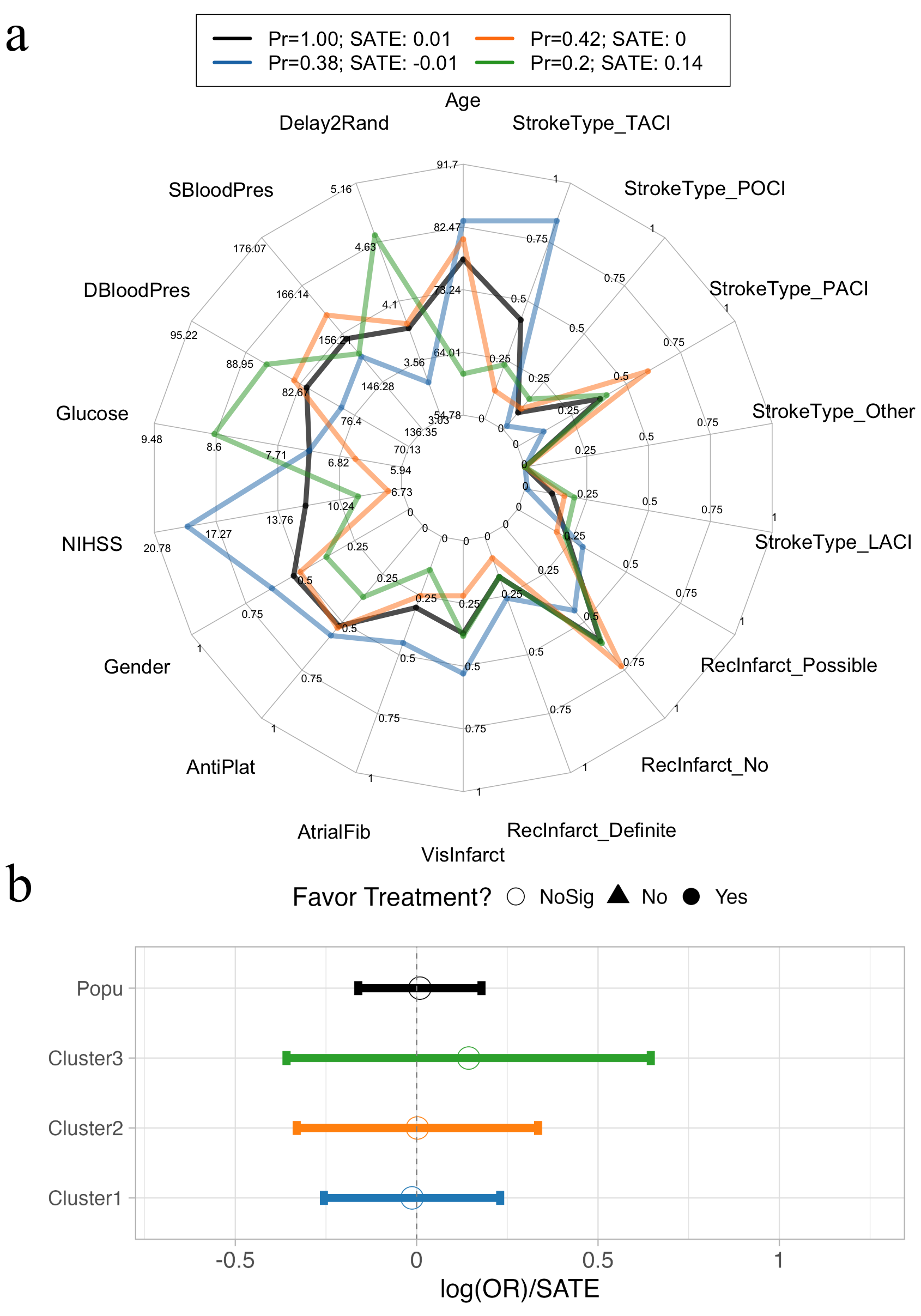}
  \caption{{\gmm} on IST-3 Dataset: \textbf{(a)} The estimated cluster means and \textbf{(b)} ORs with $95\%$ CIs (log scale) when $K=3$.}
  \label{fig:IST3 GMM res}
\end{figure}

%Resubmission 1. Clarify the model: we revised the method section to better clarify our modelling and assumptions relied on.

% 2. Baseline of supervised learning methods like causal forest, meta-learner: we included the comparison of these baseline methods in both simulation and real datasets.

\end{document}